\documentclass{egpubl}
\usepackage{egsgp2021}

 \JournalPaper         
\usepackage[T1]{fontenc}

\usepackage{cite}  

\BibtexOrBiblatex
\electronicVersion
\PrintedOrElectronic
\ifpdf \usepackage[pdftex]{graphicx} \pdfcompresslevel=9
\else \usepackage[dvips]{graphicx} \fi

\usepackage{egweblnk}

\usepackage{tikz-cd}

\graphicspath{{figures/}}
\usepackage{bbold}

\usepackage{amsfonts}
\usepackage{amsmath}
\usepackage{amssymb}
\usepackage{overpic}
\usepackage{pgfplots}
\usepackage{pgf,tikz}
\usepackage{bm}
\usepackage{wrapfig}
\usepackage{multirow}
\usepackage{mathtools}
\usepackage{enumerate}
\usepackage{verbatim}

\usepackage{tabls}

\renewcommand{\phi}{\varphi}

\definecolor{aryl}{rgb}{0.91, 0.84, 0.42}
\definecolor{aqua}{rgb}{0.7, 0.99, 0.99}

\newcommand{\R}[0]{\mathbb{R}}
\newcommand{\C}[0]{\mathbb{C}}
\newcommand{\diff}{\mathrm{d}}
\newcommand{\divg}{\mathrm{div}}
\newcommand{\I}{\mathbf{g}}

\DeclareMathOperator*{\argmin}{arg\,min}

\newcommand{\tightparagraph}[1]{\vspace{-2mm}\paragraph*{#1}}

\usepackage{soul}

\newcommand{\zo}{\textsc{ZoomOut}}

\usepackage{listings}
\usepackage{color, colortbl}
\definecolor{mygreen}{RGB}{28,172,0} 
\definecolor{mylilas}{RGB}{170,55,241}
\definecolor{myblue}{RGB}{203,192,211}

\definecolor{mycolor1}{RGB}{255,204,201}

\usepackage{booktabs}       
\newlength{\Oldarrayrulewidth}

\newtheorem{theorem}{Theorem}[section]

\newtheorem{lemma}{Lemma}[section]

\usepackage{amssymb}

\usepackage[shortlabels]{enumitem}
\usepackage{kantlipsum}

\usepackage{algorithm, algpseudocode}

\newif\ifboldnumber
\newcommand{\boldnext}{\global\boldnumbertrue}

\algrenewcommand\alglinenumber[1]{%
  \footnotesize\ifboldnumber\bfseries\fi\global\boldnumberfalse#1:}
  
\usepackage{newtxtext,newtxmath}

\title[Complex Functional Maps]%
      {Complex Functional Maps : \\a Conformal Link Between Tangent Bundles}

\author[Nicolas Donati, Etienne Corman, Simone Melzi, Maks Ovsjanikov]
{
Nicolas Donati$^{1}$, Etienne Corman$^{2}$, Simone Melzi$^{3}$, Maks
Ovsjanikov$^{1}$\\
$^{1}$LIX, Ecole Polytechnique, IP Paris, ~$^{2}$Universit\'{e} de Lorraine, CNRS, Inria, LORIA, France, ~$^{3}$Sapienza, University of Rome
}

\begin{document}
\maketitle
\begin{abstract}
  In this paper, we introduce complex functional maps, which extend
  the functional map framework to conformal maps between tangent
  vector fields on surfaces. A key property of these maps is their
  \emph{orientation awareness}. More specifically, we demonstrate that
  unlike regular functional maps that link \emph{functional spaces} of
  two manifolds, our complex functional maps establish a link between
  \emph{oriented tangent bundles}, thus permitting robust and
  efficient transfer of tangent vector fields. By first endowing and then exploiting the tangent bundle of each shape with a complex structure, the resulting operations become naturally orientation-aware, thus favoring \emph{orientation and angle preserving correspondence} across shapes, without relying on descriptors or extra regularization.
  Finally, and perhaps more importantly, we demonstrate how these
  objects enable several practical applications within
  the functional map framework. We show that functional maps and their complex counterparts can be estimated jointly to promote orientation preservation, regularizing pipelines that previously suffered from orientation-reversing symmetry errors.


\end{abstract}
\section{Introduction}
\label{sec:intro}
Non-rigid shape matching is a well-established challenge in computer graphics, geometry processing and related fields \cite{van2011survey,sahilliouglu2020recent}, with applications ranging from medical imaging to statistical shape analysis, to name a few.

One prominent direction for addressing this problem is given by the functional map framework \cite{ovsjanikov2012functional}. This framework is based on representing correspondences as linear transformations between function spaces, and encoding them as matrices using a reduced basis. A key advantage of this construction is that it allows to both optimize for and to manipulate mappings by solving small-scale optimization problems, whose complexity is largely independent of the size of the underlying meshes. Furthermore, the continuous nature of this representation enables the use of differentiable optimization techniques, which has recently proven useful in learning pipelines, e.g., \cite{FMNET,halimi2019unsupervised,ginzburg2020cyclic}.

\begin{figure}
  \vspace{-2mm}
\begin{center}
    \begin{overpic}
[trim=0.0cm 0.0cm 0.0cm 0.0cm,clip,width=1\linewidth]{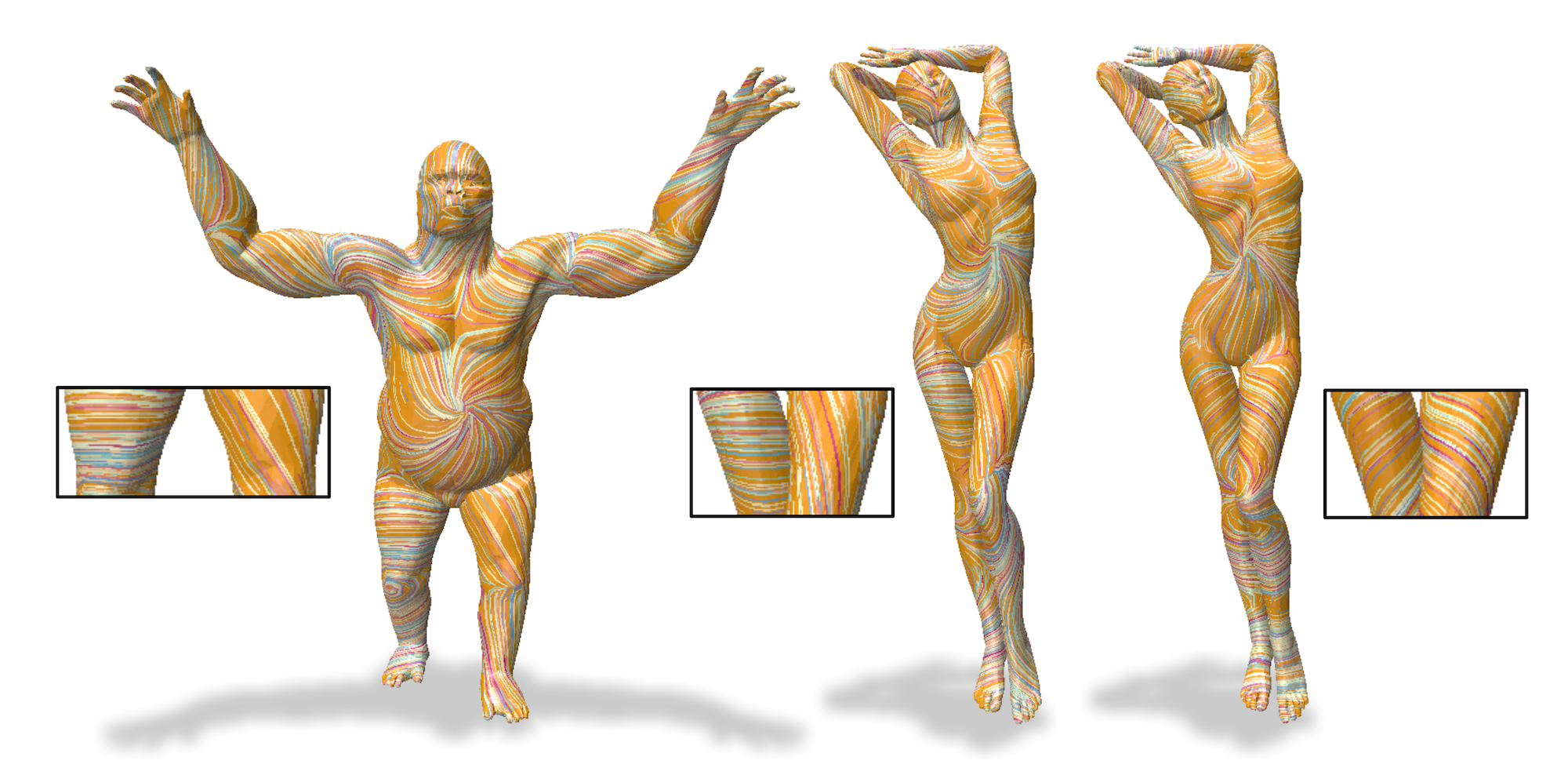}
\put(12.5,-1.5){\footnotesize Source}
\put(56,-1.5){\footnotesize Ours}
\put(73,-1.5){\footnotesize \cite{azencot2013operator}}
\end{overpic}
\end{center}
\caption{\label{fig:real_teaser} A comparison to \cite{azencot2013operator} in a vector field transfer application. The transfer is done with a functional map mixing symmetries (see Section \ref{sec:results_vf}). Unlike the method of \cite{azencot2013operator}, our approach is robust to this type of noise and transfers the vector field correctly, without breaking its asymmetry.}
\vspace{-6mm}
\end{figure}
Despite the flexibility and simplicity of the functional map representation, it has several key limitations: first, while functional maps encode correspondences between points, they do not immediately provide access to maps between derived quantities such as the surface metric or tangent vector fields that require the notion of a map differential. Several attempts have been made to recover differential information in the functional map framework, e.g., \cite{azencot2013operator,corman2017functional}. However, these approaches often lead to non-trivial optimization problems and, as we demonstrate below, can be prone to error especially when faced with approximate maps in the reduced basis (see Figure \ref{fig:real_teaser}). Perhaps even more importantly, the functional map representation does not encode information about the \emph{surface orientation}, which means that standard functional map optimization energies can easily lead to orientation-reversing correspondences that may arise, e.g., due to intrinsic symmetries. Existing methods try to tackle this challenge through a range of solutions including by using landmarks \cite{nogneng17,FARM}, injecting orientation into descriptor-based energies \cite{ren2018continuous} or alignment in the ambient space \cite{MapTree} among others. Unfortunately, these solutions often require additional user input and careful parameter tuning or incur significant computational cost.

\begin{figure}
\begin{center}
    \begin{tabular}{lr}
    \hspace{-0.3cm}

\begin{minipage}{0.5\linewidth}
    \begin{overpic}
[trim=0.0cm 0.0cm 0.0cm 0.0cm,clip,width=1\linewidth]{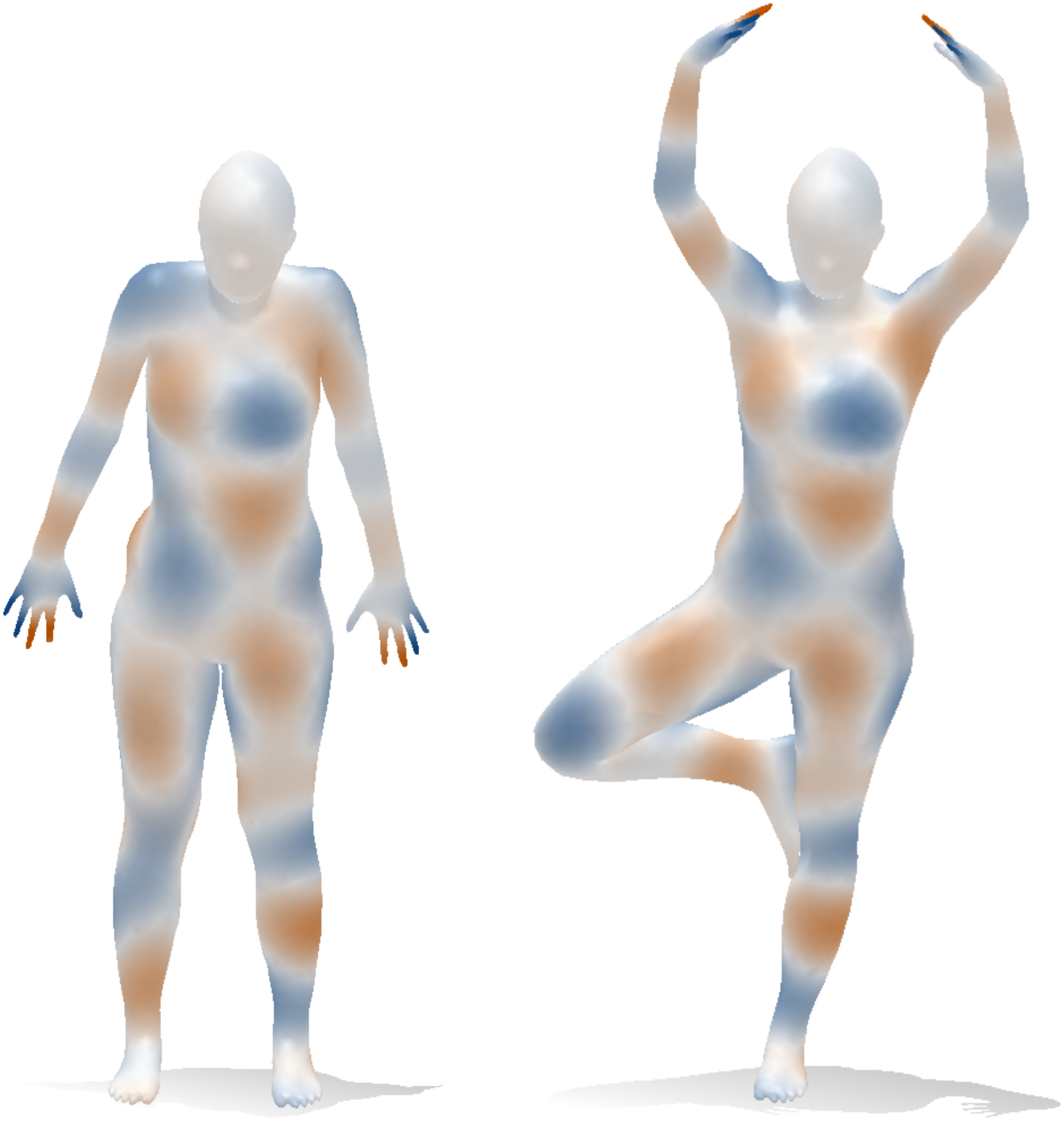}

\put(11.5,-3.9){\footnotesize Source}
\put(60,-3.9){\footnotesize Target}
\end{overpic}
\end{minipage}
    &
    \hspace{-0.50cm}

\begin{minipage}{0.5\linewidth}
    \begin{overpic}
[trim=0.0cm 0.0cm 0.0cm 0.0cm,clip,width=0.98\linewidth]{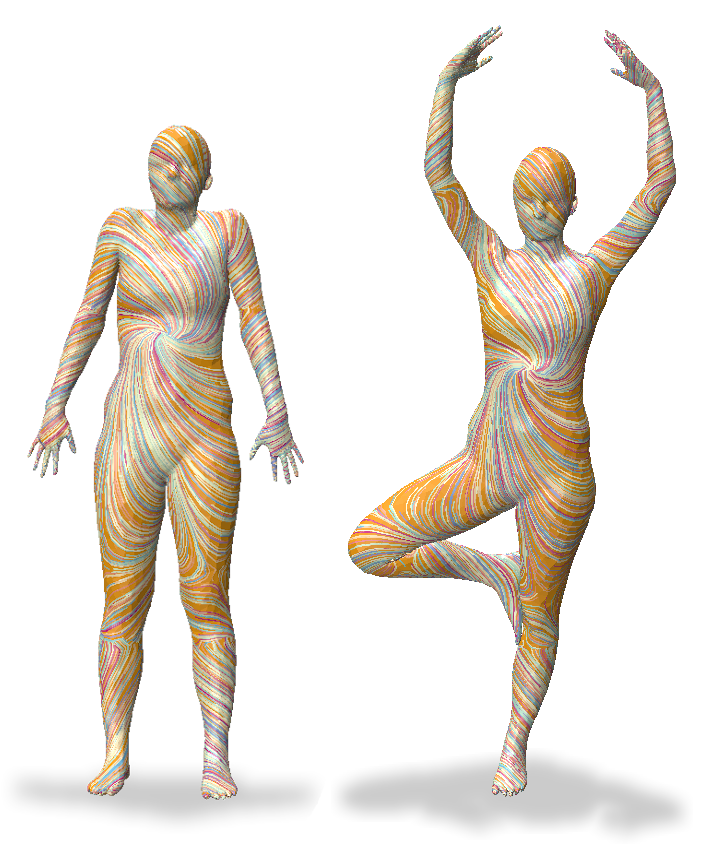}
\put(11.5,-0.5){\footnotesize Source}
\put(53,-0.5){\footnotesize Target}

\end{overpic}
\end{minipage}
\end{tabular}
\end{center}
\caption{\label{fig:teaser} A visualization of function transfer (left) and vector field transfer (right) obtained respectively via a functional map and our \emph{complex functional map}. The values of the functions on the left are encoded by a colormap, while the vector fields are visualized using line integral convolution. Both functions and vector fields can be decomposed in a Laplacian basis and be transferred using small matrices.
\vspace{-6mm}}
\end{figure}

In this work, we introduce a novel construction that helps to address these challenges directly, without relying on user input or post-processing. Our key idea is to build a representation for correspondences that \emph{only allows} orientation and angle-preserving (conformal) maps. Such a representation is, by its nature, more restrictive than the functional map representation, since only a subset of maps is allowed. However, as we demonstrate below, using this representation and especially the link with standard functional maps helps to regularize map computation and to improve accuracy in several applications without sacrificing expressive power.

To achieve this goal, we first observe that in the case of conformal maps, the pushforward (also called ``map differential") is a \emph{complex linear} operator between tangent bundles. As a result, it can be encoded as a small-sized complex-valued matrix, given a choice of basis for the tangent vector fields on each shape. Since a single vector field can be represented as \emph{a complex function}, and the pushforward allows to transfer vector fields across shapes, we call these operators \emph{complex functional maps}. As we highlight below, complex functional maps have several properties that are complementary to the standard functional map representation. Specifically, they provide a simple and robust tool for transferring vector fields; furthermore, they allow to disambiguate intrinsic symmetries and help to promote more accurate, smooth, orientation-preserving point-to-point correspondences.

\paragraph*{Contributions}
To summarize, our main contributions include:
\begin{itemize}
    \item We introduce a novel complex-linear operator acting on tangent vector fields, used as relaxation of the map differential.
    \item We demonstrate that this operator is naturally orientation-aware, and show how it can be used to regularize  functional map estimation, especially with respect to symmetry errors.
    \item We highlight the utility of our construction in a range of applications, from robust vector field transfer to orientation-aware map refinement, leading to consistent improvement in accuracy.
\end{itemize}

\section{Related Work}
Computing maps or correspondences between non-rigid 3D shapes is a key challenge in geometry processing and computer vision. Below, we review some approaches in this area and focus especially on methods that we either build upon or use as baselines, and refer to recent surveys \cite{van2011survey,biasotti2016recent,sahilliouglu2020recent} for a more in-depth discussion.

\tightparagraph{Functional maps framework} Our method heavily relies on the functional map representation, which was originally introduced in \cite{ovsjanikov2012functional} as a tool for non-rigid shape matching.
The key idea of this representation is to represent point-to-point maps as small matrices, encoded in a reduced basis, which greatly simplifies subsequent optimization problems. The original work used only a basic set of linear constraints on functional maps, which have been extended significantly in, e.g., \cite{kovnatsky2013coupled,aflalo2013spectral,huang2014functional,Eynard16,burghard2017embedding,rodola2017partial,nogneng17,huang2017adjoint,nogneng18,ren2018continuous,zoomout} among many other works (see \cite{ovsjanikov2017course}). These approaches heavily exploit the compact and continuous nature of the functional map representation and have been used to improve the accuracy, speed and robustness of the resulting shape matching pipeline. However, the functional representation itself has two major limitations that limit its applicability, as we review below.

\tightparagraph{Functional representations of differential quantities}
As mentioned in the introduction, functional maps do not naturally provide information about derived quantities such as the map differential. Several attempts have been made to recover differential information in the functional map framework. This includes operator representations for tangent vector fields \cite{azencot2013operator} and cross fields \cite{azencot2017symquadr} as well as covariant derivatives and parallel transport \cite{azencot2015discrete}. These operator representations enable tasks such as computing vector field flow efficiently, which has been used both for matching functions on surfaces \cite{azencot2016advection} and even for recovering continuous maps between shapes \cite{corman2015continuous}. Furthermore, functional representations have been also proposed for extrinsic ``deformation fields'' \cite{corman2019functional} and for capturing the intrinsic shape metric \cite{rustamov2013map,corman2017functional}, among others (see also Chapter 6 in \cite{ovsjanikov2017course} and  \cite{ben2019operator} for an overview of some recent approaches).

These constructions significantly extend the power and flexibility of the functional maps framework. At the same time, the basic concept of the map differential or pushforward, and thus reliable mapping of \emph{tangent vector fields} is still cumbersome to define and use within this formalism. As we demonstrate below, in the special case of orientation-preserving conformal maps, however, this differential has a particularly convenient representation, which provides complementary information to standard functional maps.

\tightparagraph{Orientation preservation}
Another common challenge to using the functional map representation is that it does not encode surface orientation. This implies that typically-used optimization energies, e.g., based on preservation of intrinsic descriptors, such as the HKS or WKS \cite{sun2009concise, aubry2011wave} can lead to undesirable orientation-reversing correspondences (also known as symmetry flips). One can tackle this issue by using extrinsic descriptors such as SHOT \cite{shot}, but such descriptors can be very sensitive to discretization \cite{poulenard2018multi,donati2020deep}, which can have dramatic effects on robustness. Existing methods have tried to address this challenge by using either segment \cite{ovsjanikov2012functional,kleiman2018robust} or pointwise landmarks \cite{nogneng17,FARM}, injecting orientation information with descriptor-based energies \cite{ren2018continuous}, factoring the functional space using symmetry information \cite{ovsjanikov2013shape}, alignment in the ambient space \cite{eisenberger2020smooth} or, most recently, using map space exploration strategies \cite{MapTree}. However, since the functional map representation itself does not encode orientation information, these solutions only address the problem indirectly, and, e.g., in \cite{MapTree} orientation-preserving maps are selected \emph{a posteriori} among the set of candidates, using a set of filtering criteria.

In contrast, we demonstrate that by first endowing the tangent bundle with the complex algebra using the outward normals, and then defining the derived \emph{complex functional maps} it becomes possible to directly promote orientation-preserving correspondences without any additional descriptor preservation constraints or post-processing.

\tightparagraph{Vector Field Map Representation}
Closest to our construction is the work of \cite{wang2018vector}, where the authors extend the functional map representation to differential forms on manifolds. Similarly to our approach, that method encodes the pushforward as a linear operator acting on vector fields, while imposing orthogonality, which is a necessary condition to arise from a conformal map. However, crucially, the authors of \cite{wang2018vector} use $\R$-linear operators to encode the pushforward, whereas we use the outward normals to construct and exploit the \emph{complex algebra} on the tangent space, leading to $\C$-linear operators. This difference is fundamental, as it implies that the representation in \cite{wang2018vector} \emph{cannot} distinguish orientation-preserving from orientation-reversing maps.
As we demonstrate in Section~\ref{sec:results}, this severely limits the scope of applications of that representation, which are, in contrast, enabled by our approach thanks, in particular, to its orientation-aware nature.

\section{Tangent Bundle Map as Operator}
In this section, we describe the theoretical aspects of our complex functional maps. At a high level, we follow the motivation behind the original functional map framework, by ultimately providing a linear relaxation of a particular geometric concept. While functional maps aim at representing diffeomorphisms as linear operators acting on functions, our goal is to represent the pushforward of a conformal map as a $\mathbb{C}$-linear operator acting on complex fields.

\subsection{Notation}
From now on, we consider a pair of compact Riemannian surfaces $M$, $N$ embedded in $\mathbb{R}^3$. We use $T_{p}M$ to denote the tangent plane at a point $p\in M$, while the \emph{tangent bundle} $TM := \cup_{p \in M} T_pM$, is the disjoint union of all the tangent planes of $M$. We equip this space with a proper inner product (i.e. the Riemannian metric) $\langle \cdot, \cdot \rangle_{T_{p}M}: T_{p}M \times T_{p}M \rightarrow \mathbb{R}$, which depends smoothly on $p\in M$. 
For every pair of real-valued functions $f,g:M\rightarrow\mathbb{R}$, we use another, $L^2$ inner product, which is defined as $\langle f, g \rangle_{M} = \int_M f(p)g(p)d\mu_M(p)$, where $d\mu_M$ is the area element on the surface $M$.
With respect to this inner product, we define $L^2(M)=\{ f:M\rightarrow \mathbb{R} \ s.t. \ \langle f, f \rangle_{M} < +\infty \}$, as the space of square-integrable real-valued function defined over $M$.
Similarly for every pair of vector fields $X,Y:M\rightarrow TM$ we consider their inner product $\langle X, Y \rangle_{TM} = \int_M \langle X(p), Y(p) \rangle_{T_{p}M} d\mu_M(p)$.
Finally we denote by $L_M: TM\rightarrow TM$ the connection Laplacian acting on vector fields.

\subsection{Pushforward in the smooth setting}
\begin{figure}[t!]
\begin{center}
\includegraphics[trim=0.0cm 0.0cm 0.0cm 0.0cm,clip,width=0.9\linewidth]{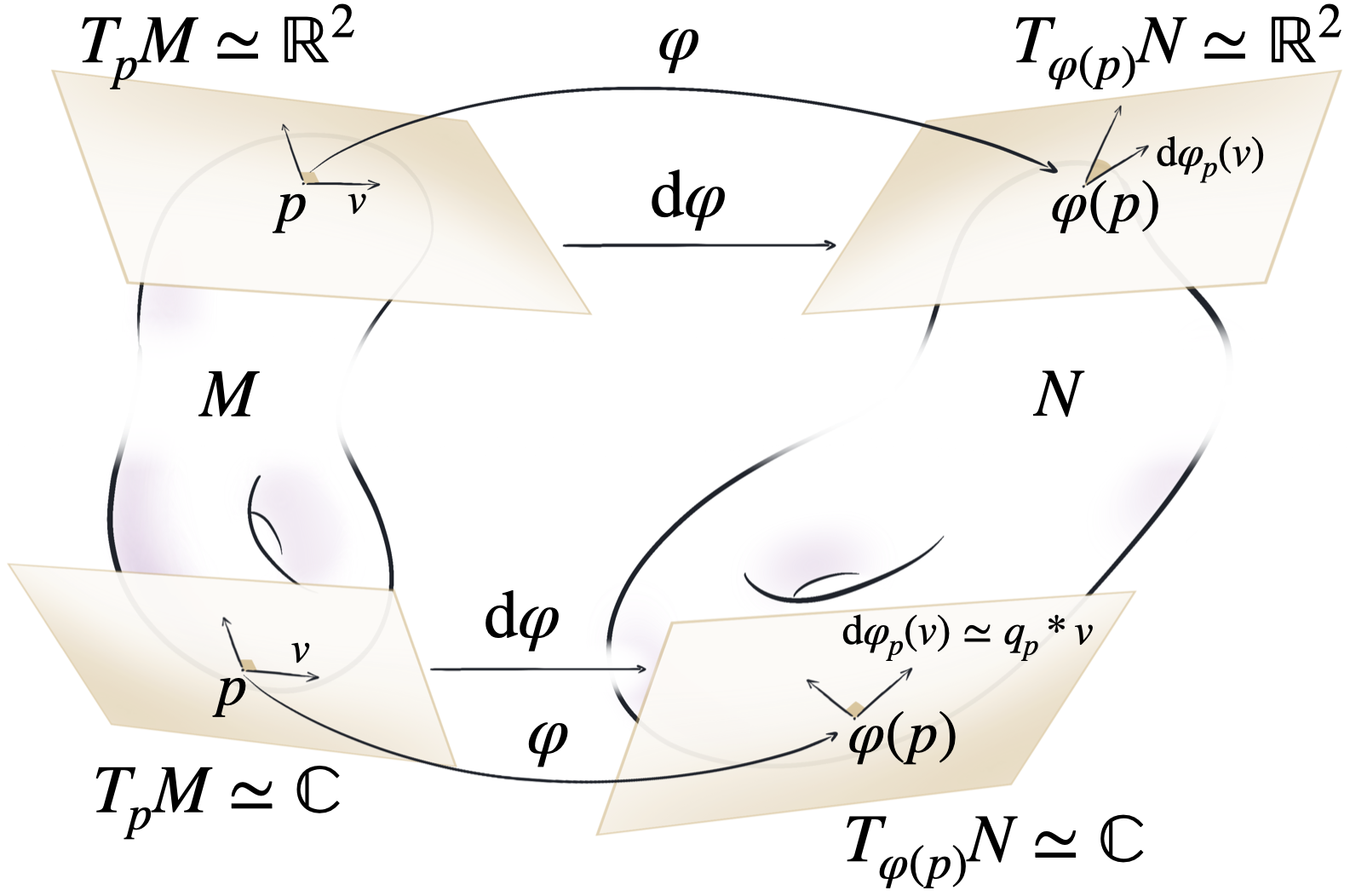}
\end{center}
\vspace{-1mm}
\caption{The pushforward $\diff \varphi$ takes a tangent vector $v \in T_pM$ and transports it to the plane tangent to $\varphi(p) \in N$. In all generality (top) surface tangent planes identify to $\R^2$ and the pushforward is locally a $\R^2$-linear map. If $\varphi$ is conformal (bottom), tangent planes identify to the complex plane $\C$ and $\diff \varphi$ is locally a multiplication by a complex number $q_p \in \C$.\vspace{-4mm}}
\label{fig:pushforward}
\end{figure}

A diffeomorphism $\varphi : M \rightarrow N$ bijectively maps points on a surface $M$ to points on a surface $N$. The pushforward $\diff \varphi : T_p M \rightarrow T_{\varphi(p)} N$ associated to $\varphi$ maps the tangent space at point $p \in M$, denoted as $T_p M$, to the tangent space $T_{\varphi(p)} N$ and can be understood as the best linear approximation of the map at the given point $p$ (see Figure~\ref{fig:pushforward} top). Thus, the pushforward contains two pieces of information: 1) which tangent plane of $N$ corresponds to a given tangent plane of $M$ and 2) \emph{how} a tangent plane is deformed by the diffeomorphism $\varphi$. While the first is already contained in $\varphi$, the second is of very different nature, and is especially difficult to recover in the discrete setting as it requires to numerically differentiate the mapping.

For surfaces, tangent planes can be identified to the vector space $\mathbb{R}^2$. Given an arbitrary diffeomorphism, the pushfoward acts \emph{linearly} on the tangent vectors (see \cite{guillemin2010differential}, Chapter 2)  so at that each point $p \in M$, $(\diff \varphi)_p$ can be represented as a linear map between two Euclidean spaces. This map can be represented by a $2 \times 2$ matrix if each space is endowed with a local 2D basis formed by two linearly independent tangent vectors. For a more in-depth discussion of map differentials, we refer to~\cite{guillemin2010differential,lee2013smooth}.

\subsection{Pushforwards of conformal maps}
In this paper, we are especially interested in the pushforward of a conformal map. Conformal maps have been widely used in computer graphics notably for texture mapping~\cite{ray2003hierarchical,springborn2008conformal}, parametrization~\cite{kharevych2006discrete,mullen2008spectral,sawhney2017boundary} and shape matching~\cite{lipman2009mobius,kim2011blended} among others. Their success is mostly due to their simple structure-preserving property: a conformal map preserves angles between tangent vectors. This means that, by definition, the pushforward of a conformal map is a \emph{similarity transformation} (i.e., a combination of rotation and uniform scaling) of the tangent space at every point.

Each tangent plane on an orientable surface can also be naturally identified with the complex plane $\C$ by identifying an arbitrary fixed direction with the real axis, and using the outward normal to determine the $90^\circ$ counter-clockwise rotations, associated with multiplication by $\imath$. Moreover, observe that for any fixed complex number $q$, the transformation $S: \C \rightarrow \C$ given by complex multiplication by $q$, and defined as $S(v) = q*v$, is a \emph{similarity transformation} of $\R^2$. To see this,  observe that in polar coordinates complex multiplication simply adds the arguments (angles) and multiplies the magnitudes. Conversely, it is easy to see that for any similarity transformation $S: \R^2 \rightarrow \R^2$,  there exists a unique complex number $q$ such that $S(v) = q*v$, where we implicitly identify $\R^2$ with $\C$ on both sides of the equality (see Figure~\ref{fig:pushforward} bottom).  These observations can be therefore summarized in the following lemma:
\begin{lemma}
\label{lemma:conformal_complex}
Given a smooth map  $\varphi : M \rightarrow N$ between surfaces, if each tangent plane $T_{p}$ and $T_{\varphi(p)}$ is identified with the complex plane, then the map $\varphi$ is conformal if and only if for every point $p\in M$ there exists a complex number $q_{p}$ such that the pushforward $\diff \varphi_p (v) = v*q_{p}$.
\end{lemma}
Stated differently, a map is conformal if and only if its pushforward, which can be thought of as a mapping between (two copies of) the complex plane, can be represented, at every point, as multiplication by some fixed complex number.

In general the pushfoward is a real-linear map between tangent spaces identified with $\R^2$. However, in the special case of \emph{conformal} mappings, the pushfoward is a $\C$-linear map. Crucially, while any $\C$-linear map is also real-linear, the converse, of course, does not hold. Remark also that up to technical conditions of differentiability, Lemma \ref{lemma:conformal_complex} is equivalent to the well-known \emph{Cauchy-Riemann equations}, which can be stated compactly as saying that a mapping is conformal if and only if the associated map differential is $\C$-linear (see \cite{remmert1991theory}, p. 51 for a discussion).

\subsection{Complex functional maps}
Following a similar reasoning as in the functional map framework \cite{ovsjanikov2012functional}, in which a functional map is \emph{any} linear transformation between functions spaces, we now consider the space of \emph{all} complex linear maps between \emph{tangent bundles}. As mentioned above, we implicitly assume that the tangent bundles are endowed with the complex structure given by some reference direction at every tangent space, and using the outward normal to define counter-clockwise rotation.

Thus we call \emph{a complex functional map} any $\C$-linear operator $Q$ that maps a complex field $X \in TM \simeq \cup_{p \in M}\C$ to a complex field $Q(X) \in TN \simeq \cup_{p \in N}\C$ on $N$, where we use $TM \simeq \cup_{p \in M}\C$ to denote the identification between tangent spaces and complex planes.

Remark that the complex structure allows us to represent any tangent vector field as a complex function, and, as a result, the operator $Q$ enables the transfer of tangent vector fields defined on $M$ to those defined on $N$. Crucially, by Lemma \ref{lemma:conformal_complex} we have that the pushforward of a conformal map must be a $\C$-linear operator and thus a complex functional map. The converse, however, does not hold, as not all $\C$-linear operators on the tangent bundle come from conformal pushforwards, in a similar way that not all linear functional mappings arise from pullbacks of pointwise correspondences.

In the remainder of the section we list the key properties of complex functional maps. In Section~\ref{sec:definition_pushforward} we exhibit a necessary and sufficient condition for a complex functional map to represent the pushforward of a conformal mapping. Section~\ref{sec:continuous_properties} studies additional useful properties, notably a weaker necessary condition for $Q$ to uniformly scale tangent planes and a characterization of isometric pushforwards as a subset of conformal maps.

\subsection{Pushforwards vs. Complex-Linear Maps}
\label{sec:definition_pushforward}
As introduced above, a complex functional map $Q : TM \rightarrow TN$ is \emph{any} $\C$-linear operator between tangent bundles. Our goal is to obtain a necessary and sufficient condition for such an operator to represent the pushforward (differential) of a conformal mapping. For this, we will use the following two key properties.

\paragraph*{Connection to pullbacks} First, suppose that $\varphi : M \rightarrow N$ is any diffeomorphism between two smooth surfaces. It is well-known that the pushforward $\diff \varphi : T_p M \rightarrow T_{\varphi(p)}N$ is the unique linear map between tangent spaces that satisfies:
\begin{equation}
    \langle X, \nabla (f \circ \varphi) \rangle_{T_pM} = \langle \diff \varphi (X), \nabla f \rangle_{T_{\varphi(p)}N},
\label{eq:defdT}
\end{equation}
for all $X \in TM, f \in L^2(N)$ (see~\cite{lee2013smooth}, Chapter 3).

%



\paragraph*{Orientability} Second, consider any $\C$-linear mapping $Q$ between tangent bundles. By linearity, for any complex number $q \in \C$ and any two complex fields $X,Y \in TM \simeq \cup_{p \in M} \C$, we have:
\begin{equation*}
	X = q Y \quad \iff \quad Q(X) = q Q(Y) .
\end{equation*}
For instance, if $q = \imath$, the map $Q$ preserves $90^\circ$ rotation. As rotations are defined relative to the local basis of the tangent plane, they carry the information of the \textit{orientation} of the manifold. Therefore \textbf{the operator $Q$ cannot change the orientation of the tangent bundle}. This property of the complex functional maps is key in our shape matching applications: it forces the maps to remain orientation-preserving. Combining these observations, we conclude that complex functional maps can only represent the differential of an \emph{orientation preserving} conformal map. Put differently, we have:

\begin{theorem}
    The complex-linear map $Q$ is a pushforward if and only if there exists an \emph{orientation-preserving} and conformal diffeomorphism $\varphi : M \rightarrow N$ satisfying:
    \begin{equation*}
        \langle X, \nabla (f \circ \varphi) \rangle_{T_pM} = \langle Q (X), \nabla f \rangle_{T_{\varphi(p)}N},
    \end{equation*}
    for all $X \in TM, f \in L^2(N)$.
\label{thm:Q_pushforward}
\end{theorem}

Thm.~\ref{thm:Q_pushforward} is a direct consequence of Lemma \ref{lemma:conformal_complex} and the uniqueness property of the map differential in Eq.~\eqref{eq:defdT}. For completeness we provide the full argument in Appendix~\ref{app:proof_pushforward}.
    


This theorem highlights the importance of Eq.~\eqref{eq:defdT} to link the complex-linear map $Q$ with the underlying pointwise map $\varphi$. Note that importantly, Eq.~\eqref{eq:defdT} only depends on the knowledge of the \emph{pullback} associated with $\varphi$, and thus provides a natural link between $Q$ and the standard functional map representation, which encodes pullbacks. In Section \ref{sec:discrete_pipeline_Q}, we show how this property can be used to relate our complex and standard functional maps in practice.

\subsection{Properties of complex functional maps}
\label{sec:continuous_properties}

In this section we provide two additional structural properties of complex functional maps.
%
%
%
%
%

\noindent{\textbf{Orthogonality.}} First, we show a \emph{necessary} condition for $Q$ to represent the differential of a conformal map: $Q$ must be an orthogonal operator. Interestingly, this is different from standard functional maps, which are orthonormal if and only if the underlying correspondence is locally area-preserving \cite{ovsjanikov2012functional,rustamov2013map}.

\begin{theorem}
    If $Q$ represents the pushforward of a conformal map $\varphi$ between surfaces, then:
    \begin{equation*}
        Q^\star Q = I ,
    \end{equation*}
    where $Q^\star$ is the adjoint operator uniquely defined by $\langle QX, Y \rangle_{TN} = \langle X, Q^\star Y \rangle_{TM}$. 
\label{thm:conformal}
\end{theorem}
 
Intuitively, this theorem comes from the fact the change of metric under the conformal map is given by some scaling factor at each tangent plane. When \emph{integrating} the inner products on the surface, this scaling factor cancels out with the change of area measure. A similar result was shown in \cite{rustamov2013map} for gradients of functions. We provide the complete proof in Appendix~\ref{app:proof_conformal}.

Importantly, Theorem \ref{thm:conformal} only establishes a necessary condition: indeed, even if $Q$ is an \emph{orthonormal} $\C$-linear operator it is only guaranteed to represent the pushfoward of a conformal map, if it satisfies Eq. \eqref{eq:defdT}.

%

\noindent{\textbf{Isometries.}} Secondly, a pushforward is isometric if and only if it commutes with the Levi-Civita connection~\cite{carmo1992riemannian} (p.181). A similar statement can be made about the connection Laplacian recently used in geometry processing~\cite{Sharp:2019:VHM} to compute transport of vector fields along geodesic paths.

\begin{theorem}
    Let $L$ be the connection Laplacian. If a complex functional map $Q : TM \rightarrow TN$ represents the pushforward $\diff \varphi$ of a conformal map, then it satisfies:
    \begin{equation*}
        L_N \circ Q = Q \circ L_M
    \end{equation*}
    if and only if $\varphi$ is an isometry.
\label{thm:isometry}
\end{theorem}

Theorem~\ref{thm:isometry}, proved in detail in Appendix~\ref{app:proof_isometry}, is very similar to that of functional maps \cite{ovsjanikov2012functional}. 




\subsection{Operators in a reduced basis}
\label{sec:reduced_basis}

In order to improve the efficiency of our algorithms it is often desirable to consider operators acting on a subspace of smooth vector fields. Since tangent planes can be identified to complex planes, we will use the \emph{complex inner product} defined as: 
\begin{align*}
    \langle X, Y \rangle_{\C_{p}M} := \langle X, Y \rangle_{T_{p}M} + \imath \langle \mathcal{J} X, Y \rangle_{T_{p}M} \in \C ,
\end{align*}
where $\mathcal{J}$ is the $90^\circ$ rotation around the normal. Note that for two vectors $x,y$ in the same complex plane this inner product is equivalent to the standard $\overline{x} y$, where $\overline{x}$ is the complex conjugate of $x$.

Suppose that $M$ is equipped with the family of vector fields $\lbrace \Psi^M_i \rbrace$ orthonormal with respect to the inner product $\langle . , . \rangle_{\C M}$. Then~\cite{reed1980methods} (Thm II.6), any vector field $X$ can be written as a linear combination:
\begin{equation*}
    X = \sum_i a_i \Psi^M_i ,
\end{equation*}
where $a_i = \left\langle X , \Psi^M_i \right\rangle_{\C M} \in \C$.
The transferred complex field $Q(X)$ can be decomposed in the basis of $N$:
\begin{align*}
    Q(X) &= \sum_j \left\langle Q(X) , \Psi^N_j \right\rangle_{\C N} \Psi^N_j \\
        &= \sum_j \Psi^N_j \sum_{i} \left\langle Q(\Psi^M_i), \Psi^N_j \right\rangle_{\C N} a_i ,
\end{align*}
where the second equality follows from the $\C$-linearity of $Q$. Thus the complex functional map can be understood as a matrix with coefficients $Q_{ji} = \left\langle \Psi^N_j , Q(\Psi^M_i) \right\rangle_{\C N}$, matching coefficients in basis on $M$ to coefficients of the basis on $N$. Figure~\ref{fig:teaser} provides a visual comparison between a classical functional map transferring functions \textit{(left)} and a complex functional map transporting tangent vector fields \textit{(right)}, both using a reduced basis of size $30$.
\section{Discrete Setting}
In this section, we introduce complex functional maps in the discrete setting. Throughout, we consider oriented manifold triangle meshes $(V,E,F)$. Our overall strategy is based on representing tangent vector fields as complex-valued pointwise functions, so that $X_i \in \C$ per vertex $i \in V$.
This choice of a point-based representation for tangent vector fields (and not face-based, as in
e.g. \cite{azencot2013operator}) is motivated by our main application: disambiguating symmetries in non-rigid shape correspondence problems. Matching vertices between surface meshes is convenient as it directly allows to transfer texture coordinates or deformations. 

We thus represent discrete complex functional maps $Q$ as complex-valued matrices mapping complex-valued pointwise
functions on $M$ to complex-valued pointwise functions on $N$. In a similar spirit to functional maps, we improve
computational efficiency by representing the operator $Q$ as a small matrix in a reduced basis of vector fields. As a
basis, we use the first eigenvectors of the \emph{connection} Laplacian, discretized as in~\cite{Sharp:2019:VHM}. 

Below we introduce Laplacian operators and the necessary local complex structure, then we construct our complex functional map and translate each continuous property in the discrete setting.

\subsection{Laplacian operators}
\label{subsec: vf_on_verts}

\setlength{\columnsep}{7pt}
\setlength{\intextsep}{1pt}
\begin{wrapfigure}[6]{r}{0.3\linewidth}
\vspace{-0.4cm}
\begin{center}
\begin{overpic}
[trim=0cm 0cm 0cm 0cm,clip,width=1.0\linewidth]{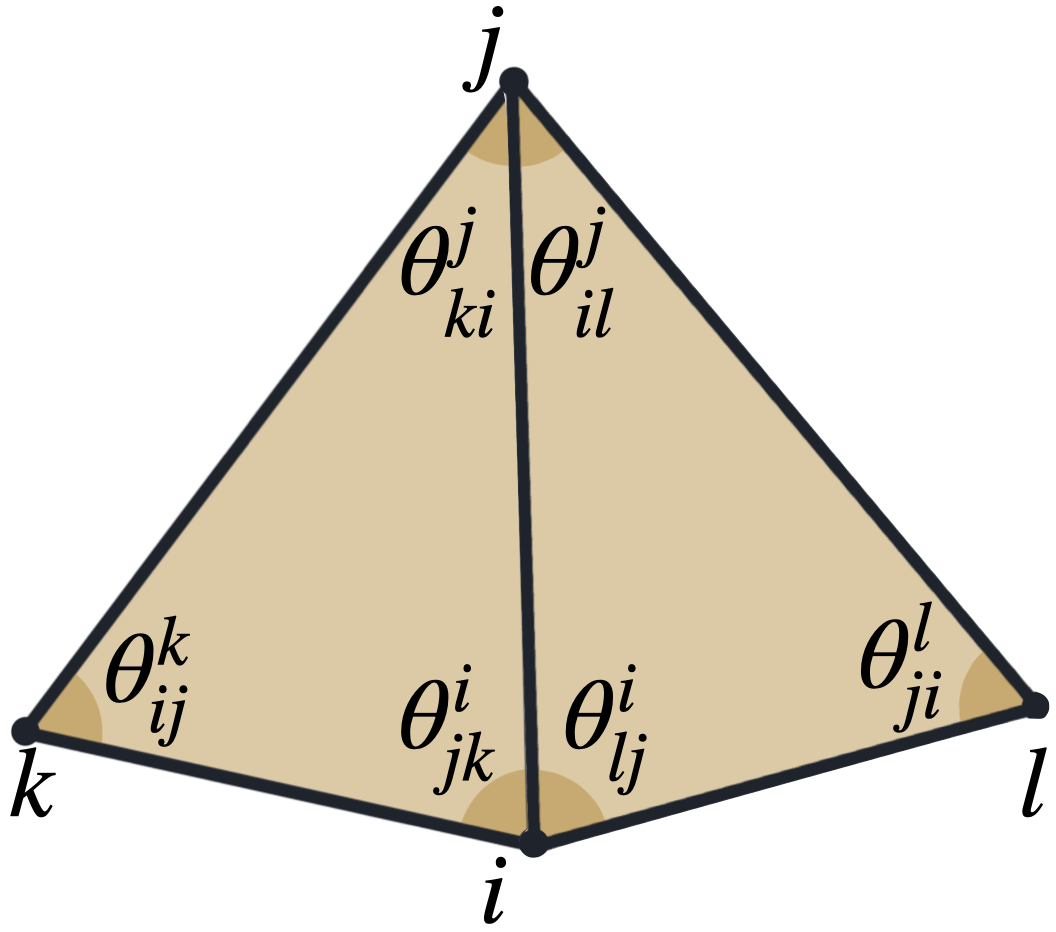}
\end{overpic}
\end{center}
\end{wrapfigure}
\paragraph*{Cotan-Laplacian}
The standard Laplacian operator $W \in \R^{|V| \times |V|}$ for a piecewise linear function $f \in \R^{|V|}$ is obtained by the well-known cotan-weight formula~\cite{botsch2010polygon}:
\begin{equation*}
    (W f)_i := \frac{1}{2} \sum_{(ij) \in E} \left( \cot \theta_{ij}^k + \cot \theta_{ji}^l \right) \left( f_i - f_j \right) ,
\end{equation*}
where the index notation is defined in the inset figure. We also define the diagonal $|V| \times |V|$ lumped mass matrix $A$:
\begin{equation*}
    A_{ii} := \frac{1}{3} \sum_{(ijk) \in T} a_{ijk} ,
\end{equation*}
where $a_{ijk}$ are the areas of triangles adjacent to $i$. This matrix defines the scalar product in the space of piecewise linear functions. Namely if $f,g \in \R^{|V|}$ are two piecewise linear real-valued functions then:
\begin{equation*}
    \langle f , g \rangle_{L^2} := f^T A g .
\end{equation*}

\paragraph*{Connection Laplacian}
Our discretization of tangent vector fields using complex numbers follows~\cite{knoppel2013globally,Sharp:2019:VHM}. 
Namely, we assign to each vertex an arbitrary reference unit vector orthogonal to the vertex normal. This reference
direction represents the tangent vector, associated with the complex number $1 + \imath 0$. The outward normal dictates
the orientation of the tangent planes, providing the additional axis $\imath$. Given this reference frame, any tangent
vector can be represented as a complex number, and a tangent vector field as a complex-valued function.  

%
%

In this context, the mass matrix $A$ defines a complex scalar product in the space of tangent fields, represented as complex functions:
\begin{equation*}
    \langle X , Y \rangle_{\C M} := X^\star A Y ,
\end{equation*}
where $^\star$ represents the conjugate-transpose operation.

As mentioned above, in our applications we use a family of smooth vector fields given by the first $k$ first eigenfunctions of the discrete \emph{connection Laplacian}, as defined in~\cite{Sharp:2019:VHM}. The connection Laplacian is the matrix $L \in \C^{|V| \times |V|}$ uniquely defined by the Dirichlet energy:
\begin{equation*}
    X^\star L X = \frac{1}{2} \sum_{(ij) \in E} (\cot \theta_{ij}^k + \cot \theta_{ji}^l) \left| X_i - r_{ij} X_j \right|^2 ,
\end{equation*}
where the unit complex number $r_{ij}$ are the rotations necessary to compare vectors expressed in two different bases.

By definition, $L$ is a complex Hermitian matrix, and similarly to the cotangent Laplacian, on Delaunay meshes has only real non-negative eigenvalues. Therefore, $L$ admits the generalized eigendecomposition:
\begin{equation*}
    L \Psi = A \Psi \Lambda ,
\end{equation*}
where $\Psi^\star A \Psi = I,$ $\Lambda$ the diagonal matrix of eigenvalues, and $\Psi$ is a set of complex-valued eigenfunctions. 
In particular, any tangent vector field $U$ can be expressed as a linear combination \emph{with complex coefficients} of
the family $\lbrace \Psi_i, i = 1, \ldots, |V| \rbrace$~\cite{horn2012matrix} (Thm.~4.1.5), \textit{i.e.}: $U = \sum_{i} c_i \Psi_i$, where $c_i \in \C$. In practice we truncate this sum and typically use a fixed number $k$ of the complex eigenfunctions, associated with the eigenvalues of smallest modulus.

\subsection{Discrete complex functional map}
\label{sec:discrete_pipeline_Q}
In order to define a discrete equivalent of the pushforward, we simply discretize the continuous definition in Eq.~\eqref{eq:defdT}. For meshes of same connectivity we are able to derive a closed-form expression for $Q$ and a consistent notion of discrete conformality. For meshes with different connectivity or when the deformation is not exactly conformal we enforce this equation in the least-squares sense.

To discretize Eq.~\eqref{eq:defdT}, we will need the pullback operator $C$, represented by a functional
map~\cite{ovsjanikov2012functional}, and the operator $D_X$ often encountered in differential geometry to define tangent
vectors~\cite{morita01} and introduced in geometry processing for vector field design in~\cite{azencot2013operator}.


To define the discrete pullback, recall that given a map $\varphi: M \rightarrow N$, the associated functional map $C:
L^2(N) \rightarrow L^2(M)$ can be discretized in the full,  ``hat'', basis as a binary matrix  $C_{NM} = \Pi_{MN}$,
where $\Pi_{MN}(i,j) = 1$ if and only if $\varphi(i) = j$, while in the reduced basis we have $C_{NM} = (\Phi^{M})^\top A^M
\Pi_{MN} \Phi^{N}$ where $\Phi^{M}, \Phi^{N}$ are matrices that store, as columns, the basis functions on the two
shapes.



\paragraph*{Vector field operator}
In addition, we will use the linear functional operator $D_X : L^2(M) \rightarrow L^2(M)$ describing the action of a vector field $X \in TM$ on a function $f \in L^2(M)$:
\begin{equation}
    D_X(f)_p := \langle X , \nabla f \rangle_{T_pM} .
\label{eq:Df_def}
\end{equation}
This operator uniquely characterizes a tangent vector field $X$ on a manifold~\cite{morita01}, and will allow us to write Eq.~\eqref{eq:defdT} as an equality between matrices.
%
Azencot \textit{et al.}~\cite{azencot2013operator} proposed a discretization for face-based vector fields, however, as
previously stated, we use a different vertex-based discretization, that we describe in detail below.

We discretize the operator $D_X \in \R^{|V| \times |V|}$ as a vertex-wise scalar product at tangent planes:
\begin{equation}
    (D_X f)_i := \langle X_i , {\nabla f}_i \rangle_{T_{i}M} ,
\end{equation}
where $\nabla f_i$ evaluates the gradient of a piece-wise linear function \emph{at vertex} $i$ (rather than at a face). In practice, we store the operator $D_X$ into a complex sparse matrix.

Locally, the gradient of a function represents the best $\R^2$-linear approximation of this function in the tangent plane. The directional derivative of $f$ along an edge vector $e_{ij}$ is simply $(f_i - f_j)/|e_{ij}|$. We therefore ask $\nabla f$ to be the best approximation of all directional derivatives, namely:
\begin{equation}
    {\nabla f}_i := \arg\min_{X \in \R^2} \sum_{(ij) \in E} \| e_{ij}^\top X - (f_i - f_j)\|^2 .
\label{eq:gradient_vertex}
\end{equation}
This least squares optimization problem amounts to pseudo-inverting a $d_i\times 2$ matrix per vertex, where $d_i$ is the degree of vertex $i$. This can be done exactly in pre-processing. 
As the gradient at vertex $i$ only depends on its neighbors, it can be encoded as a complex \emph{sparse} matrix whose non-zero coefficients per-line are equal to the degree of the vertex plus one.

\paragraph*{Discrete definition of a pusforward}
By combining these tools, we obtain a simple discretization of the definition of the pushforward in Eq.~\eqref{eq:defdT}, as a composition of functional operators, represented in the discrete setting as matrix multiplication:
\begin{equation}
    D_{X}^M C_{NM} = C_{NM} D^{N}_{QX}, \quad \forall ~X \in \C^{|V_M|} .
\label{eq:discrete_CQ_orig}
\end{equation}

Note that in this expression, consistently with Eq.~\eqref{eq:defdT}, the pushforward $Q$ maps vector fields in the
opposite direction from the pullback $C$. In practice, we have found it simpler to work with an expression that assumes that the two operators map in the same direction, which leads to:
\begin{equation}
    C_{MN} D_{X}^M  = D^{N}_{QX} C_{MN}, \quad \forall ~X \in \C^{|V_M|} .
\label{eq:discrete_CQ}
\end{equation}
This expression can be obtained simply by pre- and post multiplying Eq.~\eqref{eq:discrete_CQ_orig} by $C_{MN}$, and
assuming an invertible mapping.


In practice, mappings are usually not conformal, so we cannot hope to satisfy Eq.~\eqref{eq:discrete_CQ}
exactly. However, we can define the energy $E_{cq}$ measuring how close $Q : TM \rightarrow TN$ is  to be the differential of the operator $C_{MN} : L^2(M) \rightarrow L^2(N)$ 
and evaluating the constraint in the least squares sense:
\begin{equation}
    E_{cq}(C,Q) = \sum_i \| C D_{X_i}^M - D_{Q X_i}^N C \|^2_F,
\label{eq:energy_CQ}
\end{equation}
%
where $\{X_i\}$ is a family of vector fields in $TM$. 
Remark that Eq.~\eqref{eq:discrete_CQ} is linear with respect to $X$. Consequently, minimizing the energy of
Eq.~\eqref{eq:energy_CQ} will ensure that Eq.~\eqref{eq:discrete_CQ} is satisfied as well as possible on the subspace of
$TM$ generated by $\{X_i\}$. As above, we use the first eigenfunctions of the connection Laplacian operator for this family, and  thus the energy in Eq.~\eqref{eq:energy_CQ} ensures that Eq.~\eqref{eq:discrete_CQ} is satisfied as well as possible for \emph{smooth tangent vector fields}.

Our overall strategy thus consists in recovering $Q$ given an arbitrary functional map $C$ by solving the problem:
\begin{equation}
    \min_{Q} E_{cq}(C,Q).
\label{eq:best_Q}
\end{equation}

Eq.~\eqref{eq:best_Q} defines a simple least squares system which can be efficiently optimized by solving a linear system of equations. We describe how to do in it more details in Appendix \ref{sec:solve_Q_appendix}. The solution of this problem is the best approximation of the map differential by an \emph{orientation-preserving} conformal pushforward. The energy $E_{cq}$ is fundamental in our experiments as it allows us to extract orientation information from any given functional map.

\subsection{A closed-form expression for $Q$}
While Eq. \eqref{eq:energy_CQ} plays a fundamental role in our experiments, we remark that in the case of meshes with the same connectivity, it is possible to obtain an intuitive closed-form expression for $Q$. Recall that a conformal pushforward is given by 1) an assignment between points and 2) a similarity transformation between matching tangent planes. Therefore, it is expected that the discrete $\diff \varphi$, the pushfoward between piecewise linear complex fields, is simply the composition of a matrix $\Pi_{MN}$, assigning vertices of $M$ to those on $N$, and a multiplication by a complex field $q : N \rightarrow \C$ performing the tangent plane deformation at each vertex. Using our discrete definition of the pushforward (Eq.~\eqref{eq:discrete_CQ}), we can recover this property when the two meshes have same the connectivity.

\begin{theorem}
    Given two meshes with same connectivity and given the permutation matrix $\Pi$ describing the vertex-to-vertex
    correspondence, the solution of Eq.~\eqref{eq:best_Q} has the form:
    \begin{equation*}
        Q = D(q) \Pi ,
    \end{equation*}
    where $D(q)$ is a diagonal matrix with complex coefficients $q_i$. The complex numbers $q_i$ are the best conformal alignement of the tangent planes. 
\label{thm:discrete_Q}
\end{theorem}

The proof of Thm.~\ref{thm:discrete_Q} can be found in Appendix~\ref{app:proof_discrete_Q}.

\begin{figure}
\begin{center}
    \begin{overpic}
[trim=0.0cm 0.0cm 0.0cm 0.0cm,clip,width=1\linewidth]{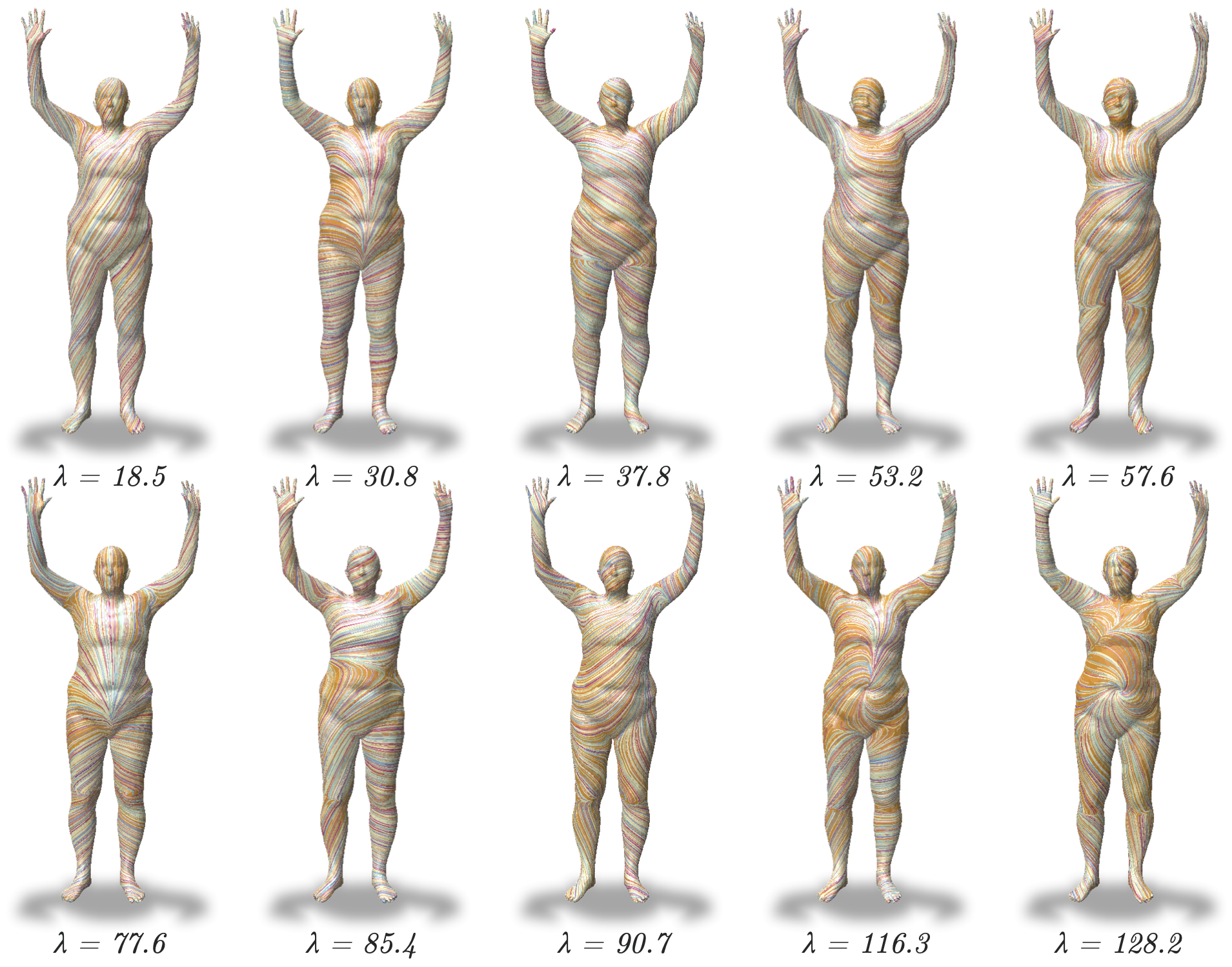}

\end{overpic}
\end{center}
\caption{LIC visualization of the first 10 eigenvectors of the connection Laplacian (from left to right, top to bottom). We see that a higher eigenvalue $\lambda$ gives an eigen tangent vector field whose frequency is higher, resulting in a less smooth flow. Nevertheless, all these tangent vector fields represent the generators of \emph{the smoothest vector fields over the shape}. It is in this basis that we will decompose tangent vector fields, as described in Section \ref{sec:Discr_reduced_basis}. 
\label{fig:vf_eigenbasis}}
\vspace{-6mm}
\end{figure}

\subsection{Constraints on complex functional maps}
\label{sec:constraints}
\paragraph*{Operator orthogonality}
While the discrete version of Thm.~\ref{thm:conformal} holds exactly in our discrete setting, it is not very informative, since the discrete notion of conformality induced by Thm.~\ref{thm:discrete_Q} is too rigid, since the only deformations of a mesh that preserve angles at every triangle exactly are isometries with possible global scaling.

However, orthonormality of $Q$ is still a valuable constraint \emph{in the reduced basis} and forces $Q$ to be the approximation of a pushforward. Furthermore, the use of a reduced basis helps to avoid both the reliance on an exact mesh structure and the rigidity of exactly conformal maps. In our experiments this constraint proved to be very effective.

In practice, we therefore solve the following Procrustes problem:
\begin{equation}
    \min_{Q^*Q=I} \; \; E_{cq}(C,Q).
\label{eq:best_Q_procrustes}
\end{equation}
This is a standard problem that can be solved exactly using a Singular Value Decomposition. For completeness we provide
the details in Appendix~\ref{sec:solve_Q_appendix}.


\paragraph*{Commutativity with the connection Laplacian}
The isometric constraint of Thm.~\ref{thm:isometry} is still valid in the discrete setting. Interestingly,
the isometry condition by enforcing the commutativity with the Laplacian is identical to the standard functional map framework. In particular, as remarked in~\cite{ovsjanikov2012functional}, it implies the more isometric a mapping is, the more diagonal our matrix $Q$ will be, when expressed in the reduced basis.

\begin{theorem}
    The conformal pushforward $Q$ commutes with the connection Laplacian:
    \begin{equation*}
        L_N Q = Q L_M ,
    \end{equation*}
    if and only if it represents an isometric map.
\label{thm:isometry_Q}
\end{theorem}

The proof is deferred to Appendix~\ref{app:proof_isometry_Q}.

\subsection{Discrete operators in a reduced basis}
\label{sec:Discr_reduced_basis}

In order to improve the efficiency of our algorithms we will not consider all piecewise linear tangent vector fields but only those spanned by a small number $k$ of smooth vector fields stored in a complex $|V| \times k$-matrix $\Psi$. 


For a known deformation across compatible meshes, the closed-form expression of $Q$ is exhibited in Thm.~\ref{thm:discrete_Q} and can easily be rewritten in a reduced basis:
\begin{equation}
    Q_{NM} = (\Psi^M)^\star A^M D(q) \Pi_{MN} \Psi^N .
\label{eq:defQ_basis}
\end{equation}

Note that Eq.~\eqref{eq:defQ_basis} is very similar to the expression of discrete \textit{functional map} $C$ introduced in~\cite{ovsjanikov2012functional}. For the function bases $\Phi^M$ and $\Phi^N$ a functional map reads:
\begin{equation}
    C_{NM} = (\Phi^M)^\top A^M \Pi_{MN} \Phi^N .
\label{eq:defC}
\end{equation}

All the constraints on $Q$ in the ``hat'' basis can be simply rewritten by replacing each term by an operator projected in the reduced basis.

In theory any orthonormal basis could be considered. For the purpose of non-rigid 3D shape matching a basis smooth and stable under nearly-isometric deformations leads to better results. In our experiments, we use the $k$ first eigenvectors of the connection Laplacian for complex functional maps, that we visualize via Line Integral Convolution in Figure \ref{fig:vf_eigenbasis}. Indeed, as proved in Thm.~\ref{thm:isometry} this operator is invariant under isometric deformations and moreover its discretization is easily implemented.


\begin{table}[t!]
\begin{center}
\begin{tabular}{|l|c|c|c|}
\hline
\multicolumn{4}{|c|}{Random noise} \\
\hline
Method / level of noise      &$s=0$&$s=0.2$&$s=0.5$\\
\hline
Wang \textit{et al.} \cite{wang2018vector}
& \textbf{0.40} & 2.2 & 5.5 \\
Azencot \textit{et al.} \cite{azencot2013operator}
& 6.2 & 4.6 & 3.2 \\
Ours & \textbf{0.40} &  \textbf{0.43}  & \textbf{0.74} \\
\hline
\hline
\multicolumn{4}{|c|}{Symmetric noise} \\
\hline
Method / level of noise      & $a=0.3$ & $a=0.5$ & $a=0.6$ \\
\hline
Wang \textit{et al.} \cite{wang2018vector}
& 0.75 & 1.1 & 1.3 \\
Azencot \textit{et al.} \cite{azencot2013operator}
& 4.6 & 1.8 & 4.2 \\
Ours & \textbf{0.40} &  \textbf{0.51}  & \textbf{0.81} \\
\hline
\end{tabular}
\end{center}
\caption{Average accuracy of the three vector field transfer algorithms on $20$ random pairs of FAUST~\cite{FAUST} for two types of noise, three noise levels. We use $k=50$ eigenvectors for both real and complex Laplacian operators. For noisy input functional maps, our method is always the most accurate. For completeness, we report more results for $k=30,70,150$ in Appendix \ref{app:VF_transfer}.
}
\vspace{-2mm}
\label{table:TVF transfer}
\end{table}

\subsection{Point-to-point map conversion}
\label{sec:Qtop2p}


In the functional map pipeline a key step is the conversion from a functional map to a vertex-to-vertex map. As originally described in \cite{ovsjanikov2012functional} and extended in \cite{pai2021fast}, one just needs to transfer Dirac functions on $M$ using the adjoint of the functional map, and compute the closest Dirac function on $N$ using a nearest neighbor search algorithm. Namely, the operation performed is $\Pi_{MN} = \mathrm{NNsearch}(\Phi^N, \Phi^M C_{NM})$, where $\Phi^M$ and $\Phi^N$ are the eigenfunctions from the standard Laplace-Beltrami operator on $M$ and $N$.

In our case, extending this algorithm to ``Dirac vector fields'' is not straightforward as a single vector is not isotropic. Instead, we propose to use the divergence operator to convert the vector field basis to functions and then use the standard functional point-to-point conversion scheme via nearest neighbor search. When $Q$ is expressed in a reduced basis, this simply amounts to computing:
\begin{equation}
    \Pi_{MN} = \mathrm{NNsearch}(\divg_N \Psi_N, \divg_M \Psi_M Q_{NM}) .
    \label{eq:Qtop2p_red}
\end{equation}

%
%
%

Where we define the discrete divergence to be the adjoint of the gradient operator matrix defined in Eq.~\eqref{eq:gradient_vertex}.
 
 This solution is not fully satisfying as it relies on the commutativity of the pushforward with the divergence operator,
and thus is geared towards near isometries. This approach, however, proves to be sufficient for our shape matching
applications in Section~\ref{sec:results}. Furthermore, in our current pipeline, $Q$ is evaluated alongside $C$ and thus, if
needed, the conversion can be done using the standard functional map. We leave finding a robust and general
conversion scheme for complex functional maps as interesting future work.

\begin{figure}
\begin{center}
    \begin{overpic}
[trim=0.0cm 0.0cm 0.0cm 0.0cm,clip,width=1\linewidth]{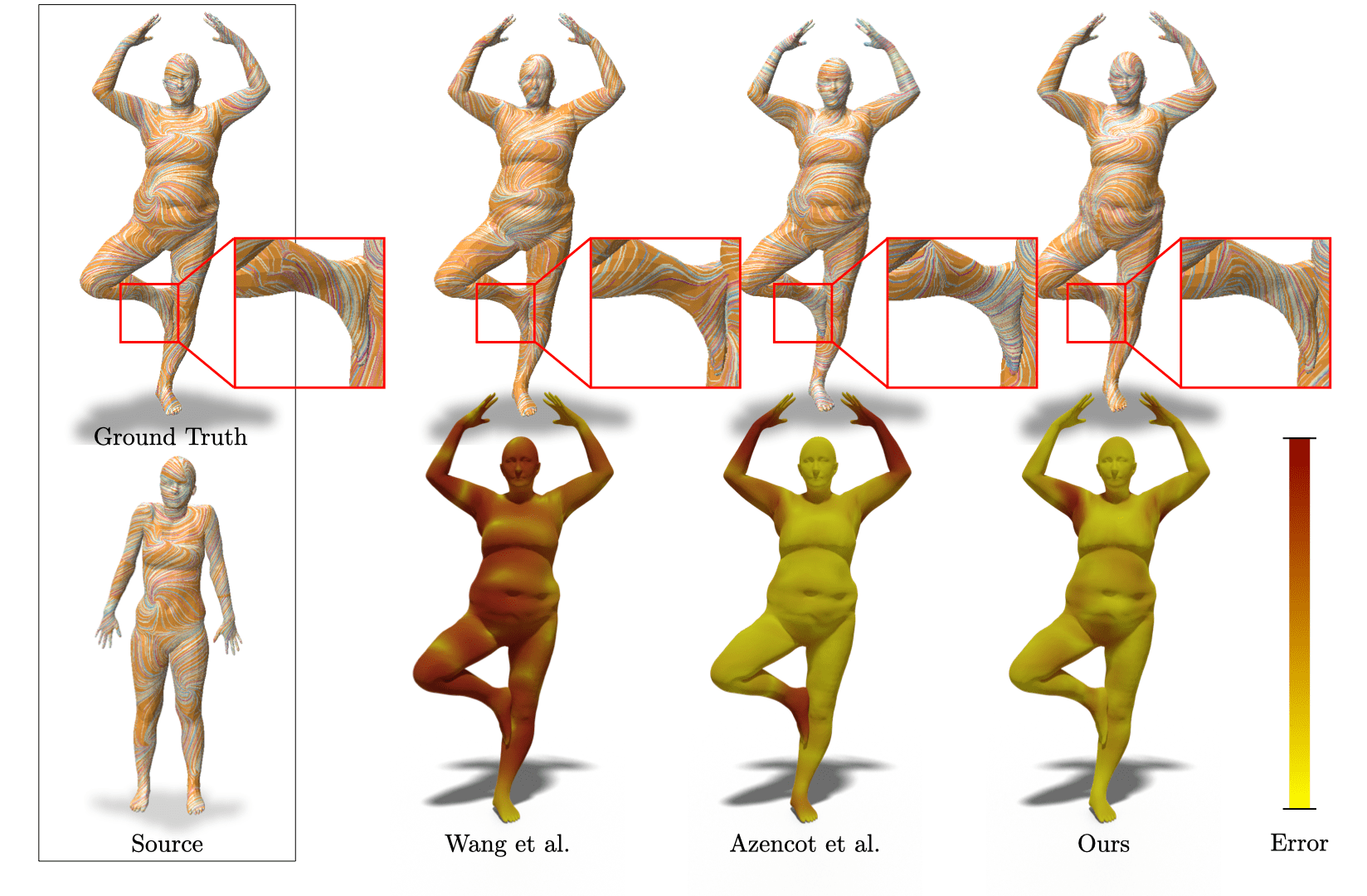}

\end{overpic}
\end{center}
\caption{Comparison of three tangent vector field transfer methods. Top row: we display a LIC visualisation of the transferred vector field on the target shape. Bottom row: we show the transfer error, computed as the difference with the ground-truth transfer. For this transfer, we used a $50 \times 50$ ground-truth functional map blurred with a random noise of magnitude $0.1$. Quantitative results can be found in Table~\ref{table:TVF transfer}.
\label{fig:VFtransfer_Error}}
\vspace{-2mm} 
\end{figure}

\section{Results}
\label{sec:results}

In this section, we present several applications of our complex functional maps. We start by demonstrating that robust vector field transfer can be achieved with complex functional maps without requiring additional information or computation.
Then, we demonstrate that the orientation-aware nature of complex functional maps can be used to eliminate the symmetry ambiguity in non-rigid near-isometric shape matching problems. Complex functional maps are easily added to standard functional map pipelines like map estimation from descriptors~\cite{ovsjanikov2012functional,nogneng17, ren2018continuous} or refinement methods like \zo \cite{zoomout} and its follow-ups \cite{MapTree, rendiscrete21}.
An implementation of our method can be found at: \href{https://github.com/nicolasdonati/QMaps}{https://github.com/nicolasdonati/QMaps}.

\begin{table}[t]
\begin{center}
\begin{tabular}{|l|c|c|}
\hline
Method (-/+ICP)  &  \cite{nogneng17}  &  \cite{ren2018continuous} \\
\hline\hline
- & 0.24 / 0.21 & 0.21 / 0.17 \\
+ Ours & 0.15 / 0.13 & 0.12 / 0.10  \\

\hline \hline
Method (-/+ICP) & \cite{wang2018vector} &  \cite{wang2018vector} + \cite{ren2018continuous}\\
\hline
- & 0.31 / 0.23 & 0.23 / 0.13 \\
+ Ours & 0.16 / 0.13 & \textbf{0.11} / \textbf{0.083} \\

\hline
\end{tabular}
\end{center}

\caption{Average geodesic error on 190 FAUST remeshed shape pairs. Our complex functional map step always improves the correspondence quality for all four algorithms, even those already incorporating the orientation-aware operators from~\cite{ren2018continuous}. Detailed graph geodesic error vs. percentage of correspondences can be found in Fig.~\ref{fig:geo curves for Table desc funmaps} of the Appendix.}
\label{table:funmap pipe with Q}
\end{table}

\subsection{Vector field transfer}
\label{sec:results_vf}

The first direct application of complex functional maps is tangent vector field transfer. We compare three methods for vector field transfer using as only input an approximate functional map $C$. We demonstrate below that using our approach is more accurate, compared to existing methods \cite{wang2018vector,azencot2013operator} especially in the presence of noise.

\paragraph*{Vector field transfer with complex functional maps}

Our representation allows for efficient and easy-to-use tangent vector field transfer. Indeed, given a functional map $C_{MN}$, we can easily recover a complex functional map $Q$ by solving the Procrustes problem in Eq.~\eqref{eq:best_Q_procrustes}. Then, transferring a vector field $X$ defined on shape $M$ to shape $N$, boils down to : 1) projecting $X$ in the complex spectral basis $X \approx \Psi^M x$ (see Section \ref{subsec: vf_on_verts} 
), 2) transferring the spectral coefficients using: $y = Q x$ and 3) recovering the output as linear combination of the target basis: $Y = \Psi^M y$.

To assess the accuracy of our method, we compare it with two standard baselines: the transfer using Hodge decomposition from Wang \textit{et al.}~\cite{wang2018vector} and the transfer using the vector field operator proposed by Azencot \textit{et al.}~\cite{azencot2013operator}.

\paragraph*{Hodge decomposition transfer}
On a shape $M$ with a sphere-topology, any tangent vector field $X \in TM$ can be decomposed as the sum of a gradient and a rotated gradient. As remarked in \cite{wang2018vector}, transferring vector fields with a conformal map can be done using only the functional map $C$ by: 1) computing the Hodge decomposition, \textit{i.e.} finding the functions $f,g \in \R^{|V|}$ such that $X = \nabla f + \imath \nabla g$, 2) transferring the functions $f,g$ using $C$ and 3) computing the gradient and rotated gradient on $N$.

%

\paragraph*{Vector field operator}
Azencot \textit{et al.}~\cite{azencot2013operator} use the
representation of a tangent vector field $X \in TM$ via the associated
functional operator $f \mapsto D_X(f)$ , defined in Eq.~\eqref{eq:Df_def} above. 
The method boils down to transferring $X$ by solving :
\begin{equation*}
    \min_Y \| D_Y C - C D_X \|^2_F .
\end{equation*}
This method is similar to ours as their energy is also inspired by Equation~\eqref{eq:defdT}. However, there are two key differences. Firstly, our method estimates the transfer for all low-frequency tangent vector fields of the source eigenbasis simultaneously by estimating $Q$, whereas the approach of Azencot \textit{et al.} is limited to one vector field transfer at a time.
Secondly, the approach in \cite{azencot2013operator} is not limited to conformal deformations, making it more flexible but also more sensitive to noisy input functional maps.

\begin{figure}[t!]
\begin{center}
    \begin{overpic}
[trim=0.0cm 0.0cm 0.0cm 0.0cm,clip,width=1\linewidth]{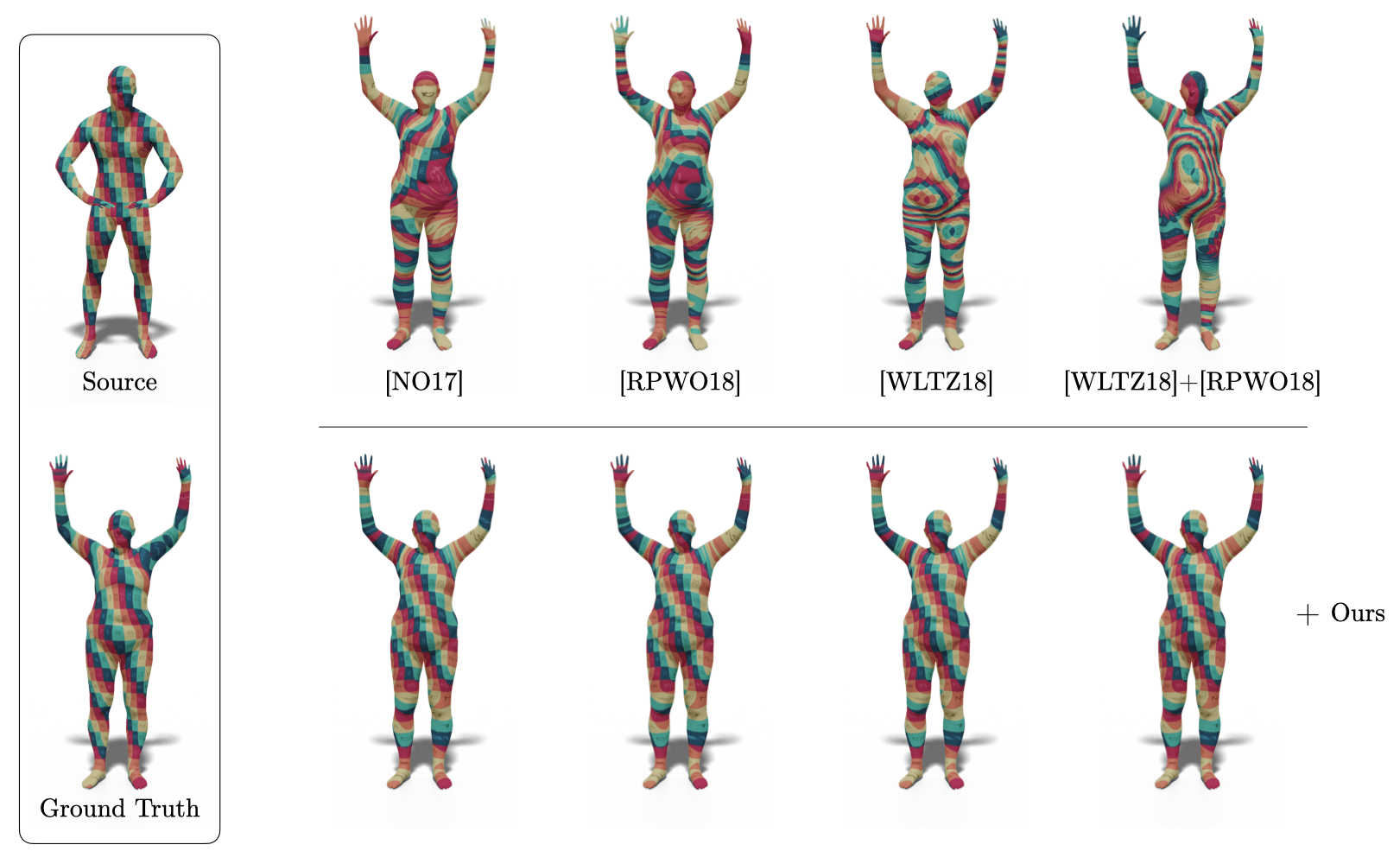}

\end{overpic}
\end{center}
\caption{We visualize the influence of complex functional map step on some descriptor-based functional map pipelines. The quality of the computed maps is displayed with texture transfer on a pair of FAUST re-meshed shapes (91 as source, 89 as target). We see that for every of the considered pipeline, our method helped with continuity and left-right symmetry errors.
Quantitative results can be found in Table~\ref{table:funmap pipe with Q}.
\label{fig:SymDisamb}}
\vspace{-2mm}
\end{figure}

\paragraph*{Results}
Table \ref{table:TVF transfer} reports a quantitative comparison between the three methods described above on $20$ random pairs taken from the original FAUST dataset~\cite{FAUST}. These meshes are in one-to-one vertex correspondence, allowing us to compute the ground-truth pushforward. 
To perform the comparison, we generate a smooth vector fields on the source, transfer it and then compute the $L^2$-distance with the exact transfer normalized by the norm of the input vector field. 
We use $k=50$ eigenvectors 
for both the functional space and the vector field space. We assess the robustness of the three functional-based vector field transfer methods by adding two types of noise to the input ground-truth $C_{gt}$ functional map:
\begin{itemize}
    \item \textbf{Random noise:} the input ground-truth $C_{gt}$ functional map is corrupted by adding a random matrix $N$ whose entries are taken uniformly at random between $-s$ and $s$, $s$ being a given threshold. We transfer tangent vector fields of the form $X = \nabla f + \imath \nabla g$ by randomizing the spectral coordinates of $f,g$ such that they decrease in intensity as frequency goes up. The results are reported in the first half of Table \ref{table:TVF transfer}. 
    \\
    \item \textbf{Symmetric noise:} The input ground-truth map $C_{gt}$ is mixed with an orientation-reversing functional map $C_{sym}$ by linearly interpolating between the two maps $C = aC_{gt} + (1-a)C_{sym}$. 
    This kind of noise often arises when estimating maps from descriptors with the original pipeline introduced in \cite{ovsjanikov2012functional}. We transfer a tangent vector field of the form $X = \nabla f$, with $f$ the extrinsic coordinate of the left-right axis. This results in an antisymmetric vector field that will only be transferred correctly if the method is robust to noise. The results are reported in the second half of Table \ref{table:TVF transfer}.
\end{itemize}

In presence of noise, our method always outperforms the baselines. This is due to the fact that we first compute the pushforward closest to $C$ in the least square sense, making it robust to random noise. By construction, our pushforward is \emph{orientation-preserving}, so it is resistant to symmetric noise and is able to recover a well-oriented transfer as shown in Fig.~\ref{fig:real_teaser}. In comparison, the other two methods directly rely on the functional map and fail if $C$ does not exactly represent a pointwise map. Moreover, if $C$ is a mix of a direct and orientation-reversing map, it is not easy to recover the underlying orientation-preserving map and thus to be robust to such noise.


In Figure \ref{fig:VFtransfer_Error} we provide qualitative illustrations of vector field transfer obtained using our method compared to baselines. Note that even a small amount of noise in the input functional map can compromise the quality of the transfer performed by \cite{wang2018vector}.
Our method only exhibits minor errors, even though it is only designed to handle conformal deformation. In fact, the only visible mis-transfer happens at the shoulder joint where the deformation is far from conformal.
The approach of Azencot \textit{et al.}~\cite{azencot2013operator} is clearly under-performing and always achieves the worst accuracy of our comparisons. However this method is more general and would give the best results in presence of perfect information \emph{and in the full basis}, even with strongly non-isometric deformations.

In conclusion, complex functional maps can help to alleviate the errors in map orientation and allow to accurately transfer vector fields between near isometric shapes even if the deformation is not exactly conformal.

\begin{table}
\begin{center}
\begin{tabular}{|l|c|c|c|}
\hline
Method / stats       & Avg. & Med. & Min. \\
\hline\hline
ZO              & 0.520 & 0.523 & 0.226  \\
ZO + Ours        & \textbf{0.320} & \textbf{0.328} & \textbf{0.037} \\
\hline\hline
ZO + bij        & 0.508 & 0.447 & 0.106 \\
ZO + bij + Ours  & \textbf{0.367} & \textbf{0.382} & \textbf{0.040} \\
\hline\hline
ZO + bij + conf  & 0.47 & 0.45 & \textbf{0.025} \\
ZO + bij + conf + Ours & \textbf{0.225} & \textbf{0.078} & 0.029 \\
\hline\hline
ZO + bij + iso  & 0.450 & 0.433 & \textbf{0.025} \\
ZO + bij + iso + Ours & \textbf{0.198} & \textbf{0.081} & 0.029 \\
\hline
\end{tabular}
\end{center}
\caption{Adding our complex functional map step in the pipeline of Ren \textit{et al.}~\cite{rendiscrete21} always improves map accuracy. We report the average, median and minimal geodesic distance error on 50 shape pairs of the SMAL dataset. Detailed graph geodesic error vs. percentage of correspondences can be found in Fig.~\ref{fig:geo curves SMAL} of the Appendix.
}
\label{table:bij ZO with Q}
\end{table}

\subsection{Disambiguating symmetry in functional maps computation}
\label{sec:results_orientation}

In our next application, we show that complex functional maps can be used within the standard descriptor-based functional map pipeline to significantly improve robustness and accuracy. The key issue that we consider is that, as remarked in prior works \cite{ovsjanikov2013shape,corman2015continuous} intrinsic descriptors \cite{sun2009concise, aubry2011wave} are often \emph{symmetric}, which can lead to poor correspondences, where a point is arbitrarily matched to the correct target point or its symmetric counterpart~\cite{MapTree}. As we demonstrate below, injecting our orientation-preserving complex maps into the pipeline can help to resolve this issue efficiently.

To achieve this, we propose to \emph{project} a given functional map into the space of orientation-preserving maps by using $Q$ as an intermediary. This projection is done in two steps. First, we approximate the associated map differential $Q$ by solving the Procrustes problem in Eq.~\eqref{eq:best_Q_procrustes}. Secondly, we extract from $Q$ the underlying point-to-point mapping using the algorithm described in Section~\ref{sec:Qtop2p}.
Since, by construction, $Q$ is orientation preserving, the projection removes the orientation reversing component of the input map, and thus the resulting point-to-point mapping should be orientation preserving.

\begin{figure}
\begin{center}
    \begin{overpic}
[trim=0.0cm 0.0cm 0.0cm 0.0cm,clip,width=1\linewidth]{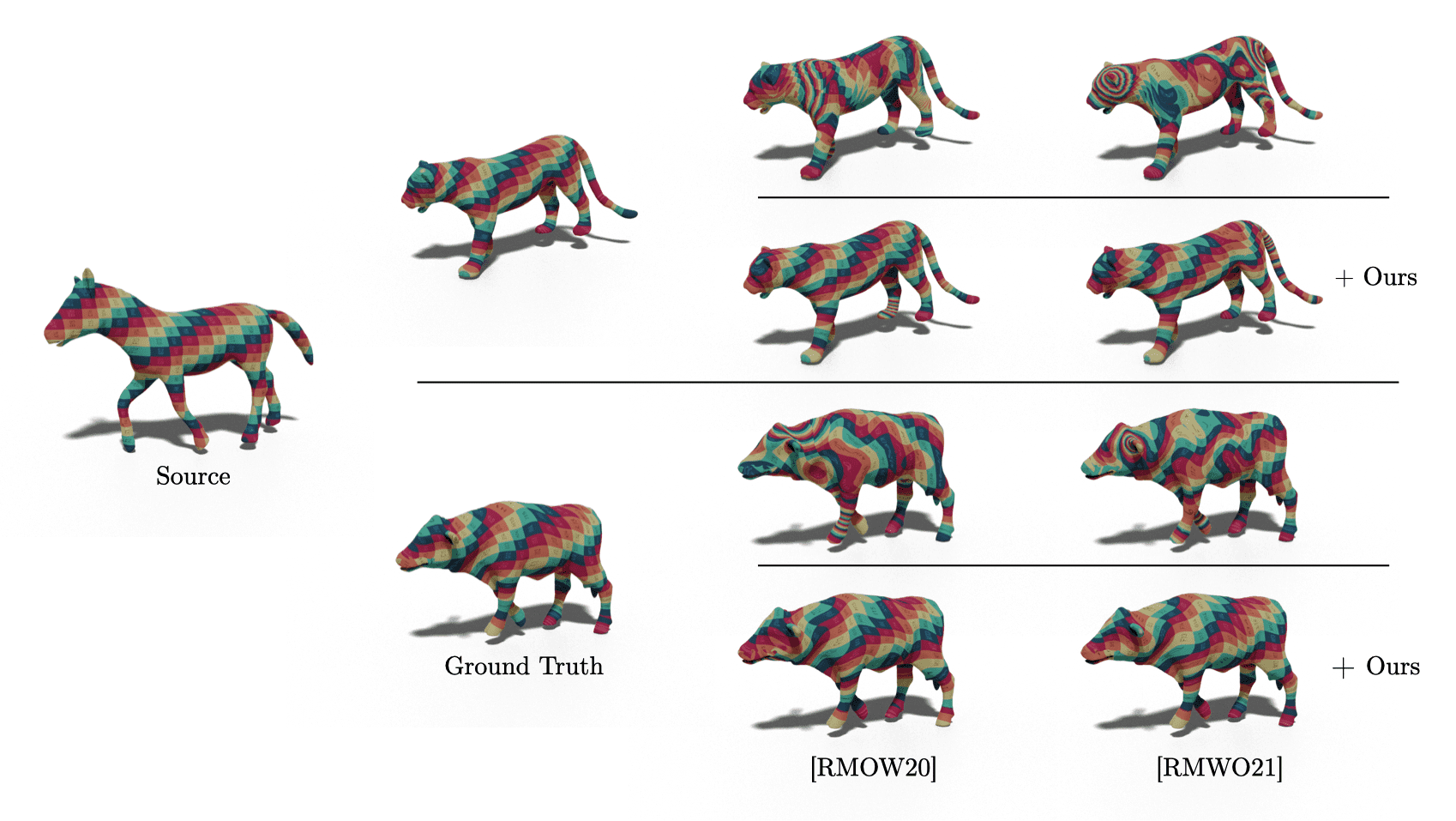}

\end{overpic}
\end{center}
\caption{Texture transfer on SMAL re-meshed dataset \cite{Zuffi2017} illustrating maps obtained with two methods based on \zo: bijective \zo \cite{MapTree} (third column) and bijective \zo{ }with isometric energy \cite{rendiscrete21} (last column). Adding our $Q$-step (bottom sub-rows) considerably improves the accuracy of the map. For quantitative results see Table~\ref{table:bij ZO with Q}.
\label{fig:bijZOQ}}
\vspace{-5mm}
\end{figure}

In a third optional step, one can reconstruct a new functional map from the point-to-point map using Eq.~\eqref{eq:defC}. This allows us to improve the mapping using post-processing technique like the spectral ICP refinement introduced in~\cite{ovsjanikov2012functional}.

We compare this approach to four alternative algorithms for computing functional maps solely from $50$ WKS~\cite{aubry2011wave} descriptors:
\begin{itemize}
    \item The algorithm introduced by Nogneng \textit{et al.}~\cite{nogneng17} where descriptors are converted into operators that must commute with the functional map.
    \item We add the orientation-promoting operators of~\cite{ren2018continuous} to the pipeline of Nogneng \textit{et al.}~\cite{nogneng17}.
    \item The method proposed in \cite{wang2018vector}, in theory closest to ours, which proposes to also transfer differential 1-forms with functional maps. It differs from our method mainly because they rely on $\R^2-linearity$ whereas our maps are $\C$-linear, and thus orientation-aware. In \cite{wang2018vector} the authors transform descriptor functions into tangent vector fields by taking the gradient.
    \item A combination of \cite{wang2018vector} with the orientation-aware operators of~\cite{ren2018continuous}.
\end{itemize}
To assess the ability of $Q$ to recover orientation information, we compare maps obtained by direct conversion of the functional map computed using each of approaches above, with maps obtained by conversion from $Q$, that has first been estimated from $C$.

The results are computed on 20 shapes (which corresponds to 190 pairs) from the FAUST dataset~\cite{FAUST}, which were re-meshed in \cite{ren2018continuous} in order to remove the bias of identical triangulations. We then compare the mean geodesic distance error obtained by these different methods -- for each source point, we compute the geodesic distance on the target between the point mapped by the obtained maps and the point mapped by ground-truth. The results are reported in Table~\ref{table:funmap pipe with Q}. Additionally, we display in Figure \ref{fig:SymDisamb} a qualitative result where our additional step was able to disambiguate symmetry whereas the standard pipelines could not.



This experiment shows that our $Q$-map projection step significantly improves the accuracy of the correspondence \emph{with all algorithms}, even in the presence of orientation-aware operators.

\subsection{Orientation-preserving \zo}
\label{sec:or_pres_zo}

Finally, our complex functional maps can also be used to improve more recent functional map algorithms based on spectral upsampling, inspired by \zo \cite{zoomout}. The \zo{ }algorithm starts with an initial pointwise map $\Pi_0$ and alternates between two steps: estimating a new functional map $C_n$ from $\Pi_n$ and recovering the new pointwise map $\Pi_{n+1}$ from $C_n$. In order to increase the map precision, the size of the spectral basis increases at each iteration. 


Interestingly, we can modify the basic \zo{ }approach to incorporate complex functional maps and thus promote discovery of orientation preserving maps. Our new algorithm basically boils down to 3 steps instead of 2: 1) estimate the functional map $C_n$ from $\Pi_n$. \emph{2) \textbf{[$Q$-step]} Estimate the complex functional map $Q_n$ from $C_n$, using Eq.~\eqref{eq:best_Q_procrustes}.} 3) Estimate the new pointwise map $\Pi_{n+1}$ \emph{from $Q_n$} (instead of $C_n$ like in classic \zo) using Eq.~\eqref{eq:Qtop2p_red}. The pseudo-code can be found in Algorithm \ref{alg:ZO_Q_orig} where the lines that describe our modification are highlighted in bold. 

We remark that this ``$Q$-step" can easily be added to other algorithms built on top of \zo. MapTree~\cite{MapTree}, which uses a tree structure to explore the space of maps, modifies \zo{ }to promote bijectivity. This idea was later extended to other properties like conformality or isometry by Ren \textit{et al.}~\cite{rendiscrete21}. Like \zo, these algorithms are based on spectral upscaling and conversions between spectral mappings and vertex-to-vertex maps. We describe their modification in Appendix~\ref{app:algo}.

\begin{algorithm}[t!]
\caption{Complex \zo}
 \label{alg:ZO_Q_orig}
\begin{algorithmic}[1]
\State \textbf{Intput:} Manifold meshes $M$ and $N$
\State Initial pointwise maps $\Pi_{MN}$
\State \textbf{Output:} Refined maps $\Pi^{ref}_{MN}$
\State \textbf{Parameters:} The number of refinement steps $J$
\State An array $[k_j], j \in [1,J]$ with the (increasing) number of spectral coordinates to use at each refinement step
\State \textbf{Preprocessing:} Compute the Laplace Beltrami eigenbases $\Phi^M$ and $\Phi^N$ (used for function in spectral basis)

\boldnext
\State Compute the connection Laplacian complex eigenbases $\Psi^M, \Psi^N$ (used for vector field spectral bases, see Section \ref{sec:Discr_reduced_basis})
\boldnext
\State Compute the differential operators $D_{\Phi^M_i},$ and $D_{\Phi^N_i}$ for $i \in [1, k_J]$ (used for estimating $Q$ from $C$, see Section \ref{sec:discrete_pipeline_Q})
\boldnext
\State Compute the reduced divergence operators $\divg_M$ and $\divg_N$ (used for conversion from $Q$ to pointwise map, see Section \ref{sec:Qtop2p})
 \For{$k \in [k_1, ..., k_J]$}
    \State $\Phi_M = \Phi^M_{[1,k]}, \Phi_N = \Phi^N_{[1,k]}$

    \State $C_{NM} = \Phi_M^\dagger \Pi_{MN} \Phi_N$

    \boldnext
    \State $Q_{NM} = \argmin_{Q \in \mathcal{O}(k)} \sum_{i=1}^k \|C_{NM} D_{\Psi^N_i} - D_{Q \Psi^N_i} C_{NM}\|_F^2$

    \boldnext
    \State $\Pi_{MN} = \text{NNsearch}(\divg_N \Psi_N,  \divg_M \Psi_M Q_{NM})$

\EndFor

\end{algorithmic}
\end{algorithm}


We demonstrate the beneficial effect of this $Q$-step by refining
\emph{random} functional maps with four versions of \zo{ }with and
without our modification. We upscale the maps from 4 eigenfunctions up
to 50, with a step of 1, and 10 inner loops per step. We perform this
experiment on 50 shape pairs of SMAL~\cite{Zuffi2017} re-meshed. We
report the obtained quantitative results in Table~\ref{table:bij ZO
  with Q}. For the baselines, we used the standard \zo{ }algorithm
\cite{zoomout}, bijective \zo \cite{MapTree}, and the discrete
optimization with first conformal \textit{(conf)} and isometric energy
\textit{(iso)} \cite{rendiscrete21}. For all of these baselines, we
report both their overall geodesic error (mean, median and minimum
error) and that of their modification with our method. We observe that
our modification is always relevant in this case, resulting in a
significant boost in overall accuracy on this dataset. More
specifically, the discrete optimization approach \cite{rendiscrete21}
with complex functional maps performs really well (median error below
$0.09$) despite the fact it is randomly initialized, and does not use \emph{any} descriptors.


In Figure \ref{fig:bijZOQ} we provide qualitative results with two shape pairs with the same source, and report five maps per pair: the ground-truth, bijective \zo \cite{MapTree}, and bijective isometric \zo \cite{rendiscrete21}, as well as their version with our modification. For both shape pairs and both methods, the original algorithms are affected by the left-right symmetry and converge to discontinuous maps. 

In summary, in state-of-the-art refinement pipelines involving \zo, our $Q$-step appears to promote orientation preservation and continuity, resulting in better convergence and more accurate results overall. This confirms that using complex functional maps in functional map pipelines is beneficial to estimate high quality correspondences from very low-frequency or extremely unreliable initialization.


\section{Conclusion, Limitations \& Future Work}
In this paper, we have introduced a new functional operator resulting from a $\C$-linear relaxation of the space of pushforwards: the complex functional map. This operator allows to robustly transfer tangent vector fields between non-rigid surfaces in 3D. Furthermore, the most prominent property of this new tool is that it reflects the complex structure of the surfaces and is thus orientation-aware. In our experiments, we exploited the orientation-aware property of complex functional maps in several shape matching tasks.
This contribution to the functional framework considerably increases the applicability of intrinsic shape matching methods, which are often hindered by the presence of orientation-reversing intrinsic symmetries. This is particularly relevant for functional map-based algorithms, since they rely on a linear relaxation of diffeomorphisms, which can linearly blend direct and orientation-reversing maps, potentially resulting in large discontinuities.

However our framework is naturally restricted to differentials of \emph{conformal} orientation-preserving mappings. This constraint is only enforced in a least squares sense in our approach and thus helps to \emph{promote} conformality which can be beneficial for shape pairs satisfying this assumption. Besides, like most algorithms derived from the functional framework, our construction is dependent on the choice of reduced basis. This can limit its applicability in more general non-isometric shape matching. As follow-up work, we would like to study other representations for orientation-preserving pushforwards and alleviate the dependency on the reduced basis.

In the future it would also be interesting to investigate the utility of complex functional maps in other applications that involve vector field transfer, including deformation or pose transfer, or synchronized convolution in the context of geometric deep learning. Indeed, restricting the search to well-oriented maps without additional supervision could lead to new efficient methods in unsupervised 3D deep learning.

\paragraph*{Acknowledgements}
The authors would like to thank the anonymous reviewers for their helpful feedback and suggestions. Parts of this work were supported by the ERC Starting Grants No. 758800 (EXPROTEA) and No. 802554 (SPECGEO), and the ANR AI Chair AIGRETTE.

\bibliographystyle{eg-alpha-doi}
\bibliography{egbibsample}
\appendix
\section{Proof of Theorem~\ref{thm:Q_pushforward}}
\label{app:proof_pushforward}
    
    \textit{Necessary condition:} Let $Q$ be the differential of an orientation-preserving conformal diffeomorphism $\varphi$. Then by Lemma \ref{lemma:conformal_complex} $Q$ must be $\C$ linear and by virtue of being a differential, $Q = \diff \varphi$ must satisfy Eq.~\eqref{eq:defdT},~\cite{lee2013smooth}.

    \textit{Sufficient condition:} Let $Q$ be a $\C$ linear operator and suppose that there exists a diffeomorphism $\varphi : M \rightarrow N$ such that $Q$ satisfies Eq.~\eqref{eq:defdT}. Since the pushforward $\diff \varphi : TM \rightarrow TN$ is the \emph{unique} operator satisfying Eq.~\eqref{eq:defdT} (see~\cite{lee2013smooth}, Chapter 3), we must have $Q = \diff \varphi$.
    %
    Now, since $Q$ is a $\C$-linear pushforward, it preserves both the angles between the vectors and the orientation of the tangent bundle. Therefore, the map $\varphi$ must then be an orientation-preserving conformal map.

\section{Proof of Theorem~\ref{thm:conformal}}
\label{app:proof_conformal}
  Let $\varphi : M \rightarrow N$ be a conformal diffeomorphism and $Q = \diff \varphi : TM \rightarrow TN$ its corresponding pushforward. By definition of conformality there exists a log-scale factor $u : M \rightarrow \R$ relating the metric tensors of both surfaces $\varphi^\star \I^N = e^{2u} \I^M$ and their volume form $\varphi^\star \diff \mu_N = e^{2u} \diff \mu_M$.
    Thus, the $L^2$ scalar product between vector fields $X,Y \in TM$ is preserved by the pushforward:
    \begin{align*}
        \int_{N} \I^N_p \left( Q(X), Q(Y) \right) \diff\mu_N(p) &= \int_{N} \I^N_p \left( \diff \varphi(X), \diff \varphi(Y) \right) \diff\mu_N(p) \\
            &= \int_M \I^N_{\varphi(q)} \left( \diff \varphi(X), \diff \varphi(Y) \right) e^{-2u} \diff\mu_M(q) \\
            &= \int_M (\varphi^\star \I^N)_q \left( X, Y \right) e^{-2u} \diff\mu_M(q) \\
            &= \int_M \I^M_q \left( X, Y \right) \diff\mu_M(q)
    \end{align*}
    This this holds for arbitrary $X,Y$, we obtain that $Q^\star Q = I$.

\begin{algorithm}[t!]
\caption{Complex Bijective \zo}
 \label{alg:bij_ZO_Q}
\begin{algorithmic}[1]
\State \textbf{Intput:} Manifold meshes $M$ and $N$
\State Initial pointwise maps $\Pi_{MN}$ and $\Pi_{NM}$
\State \textbf{Output:} Refined maps $\Pi^{ref}_{MN}$ and $\Pi^{ref}_{NM}$
\State \textbf{Parameters:} The number of refinement steps $J$ 
\State An array $[k_j], j \in [1,J]$ with the (increasing) number of spectral coordinates to use at each refinement step
\State \textbf{Preprocessing:} Compute the Laplace Beltrami eigenbases $\Phi^M$ and $\Phi^N$ (used for function in spectral basis)

\boldnext
\State Compute the connection Laplacian complex eigenbases $\Psi^M$ and $\Psi^N$ (used for vector fields in spectral basis, see Section \ref{sec:Discr_reduced_basis})
\boldnext
\State Compute the differential operators $D_{\Phi^M_i}$ and $D_{\Phi^N_i}$ for $i \in [1, k_J]$ (used for estimating $Q$ from $C$, see Section \ref{sec:discrete_pipeline_Q})
\boldnext
\State Compute the reduced divergence Operators $\divg_M$ and $\divg_N$ (used for conversion from $Q$ to pointwise map, see Section \ref{sec:Qtop2p})
 \For{$k \in [k_1, ..., k_J]$}
    \State $\Phi_M = \Phi^M_{[1,k]}, \Phi_N = \Phi^N_{[1,k]}$
    \State $C_{MN} = \Phi_N^\dagger \Pi_{NM} \Phi_M, \; \; C_{NM} = \Phi_M^\dagger \Pi_{MN} \Phi_N$

    \boldnext
    \State $Q_{MN} = \argmin_{Q \in \mathcal{O}(k)} \sum_{i=1}^k \|C_{MN} D_{\Psi^M_i} - D_{Q \Psi^M_i} C_{MN}\|_F^2$
    \boldnext
    \State $Q_{NM} = \argmin_{Q \in \mathcal{O}(k)} \sum_{i=1}^k \|C_{NM} D_{\Psi^N_i} - D_{Q \Psi^N_i} C_{NM}\|_F^2$

    \boldnext
    \State $\Pi_{MN} = \text{NNsearch}(\divg_N \Psi_N, \divg_M \Psi_M Q_{NM})$
    \boldnext
    \State $\Pi_{NM} = \text{NNsearch}(\divg_M \Psi_M, \divg_N \Psi_N Q_{MN})$
    
    \State 
    $C_{MN} = \begin{pmatrix} \Phi_N \\ \Pi_{MN} \Phi_N \end{pmatrix}^\dagger \begin{pmatrix} \Pi_{NM} \Phi_M \\ \Phi_M \end{pmatrix}$
    \State 
    $C_{NM} = \begin{pmatrix} \Phi_M \\ \Pi_{NM} \Phi_M \end{pmatrix}^\dagger \begin{pmatrix} \Pi_{MN} \Phi_N \\ \Phi_N \end{pmatrix}$
    
    \State $\Pi_{MN} = \text{NNsearch}( \big( \Phi_N C_{NM} \; \; \Phi_N C_{MN} \big), \big( \Phi_M \; \; \Phi_M \big) )$
    \State $\Pi_{NM} = \text{NNsearch}( \big( \Phi_M C_{MN} \; \; \Phi_M C_{NM} \big), \big( \Phi_N \; \; \Phi_N \big) ) $
    
\EndFor
 
\end{algorithmic}
\end{algorithm}

\section{Proof of Theorem~\ref{thm:isometry}}
\label{app:proof_isometry}
    Let us assume that $L_{M} = Q^{-1} \circ L_N \circ Q$. As shown in \cite{berline2003heat,Sharp:2019:VHM}, the diffusion kernel of the connection Laplacian is, at first order, the parallel transport along a geodesic and its magnitude is identical to the decay of the scalar heat kernel. Since the pushforward preserves the connection Laplacian, it also preserves the scalar heat kernel, therefore it must be an isometry~\cite{sun2009concise}.

    As proved in~\cite{carmo1992riemannian} (p.181) the pushforward of the Levi-Civita connection by a conformal mapping is itself the Levi-Civita connection if and only if the map is an isometry. Therefore, if $\varphi$ is an isometry, the pushfoward of the connection Laplacian $Q^{-1} \circ L_N \circ Q$ is equal to the connection Laplacian on $M$. 
    
\begin{figure}
\begin{center}
    \begin{overpic}
[trim=0.0cm 0.0cm 0.0cm 0.0cm,clip,width=1\linewidth]{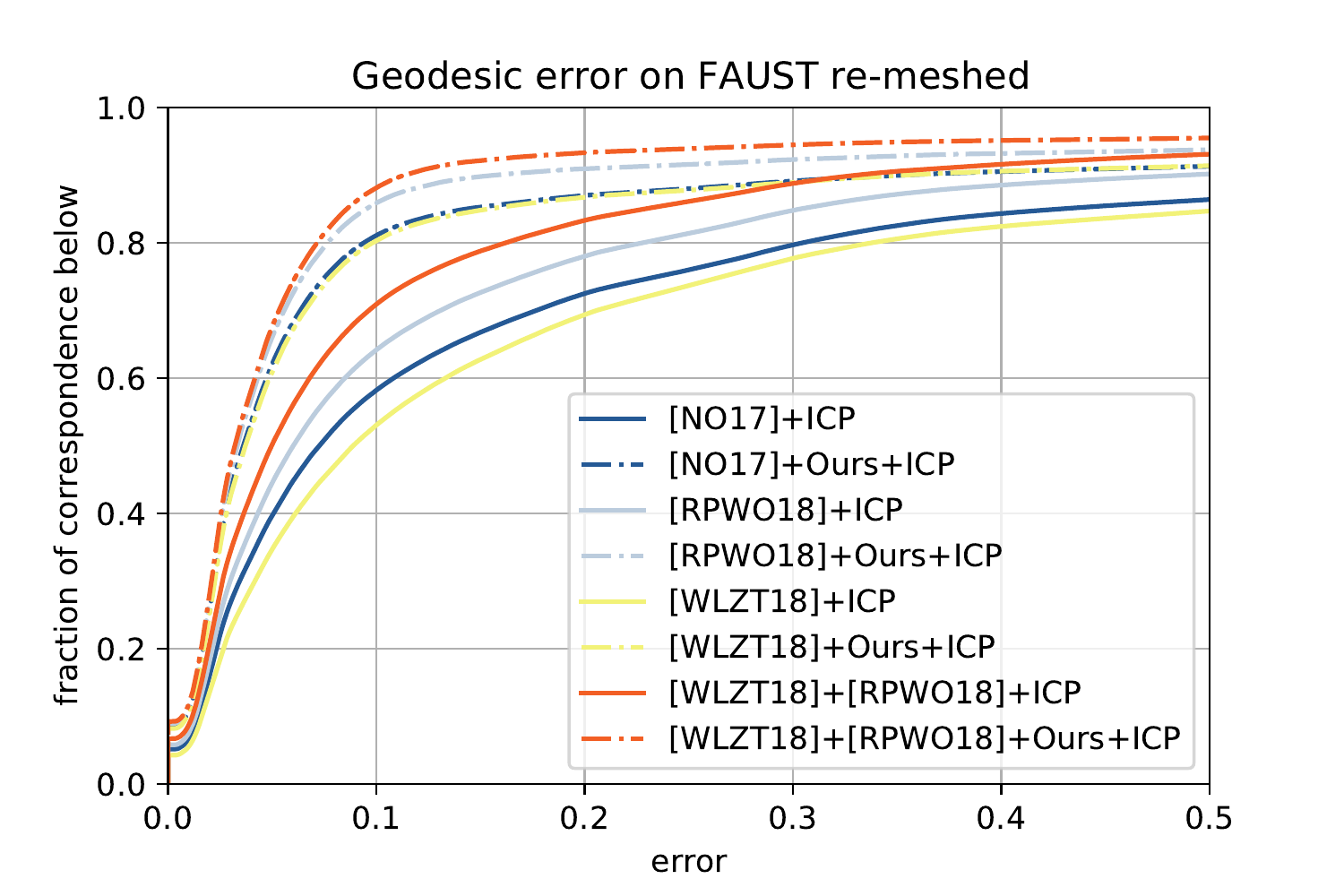}
\end{overpic}
\end{center}
\caption{Geodesic error of different methods with and without our $Q$-step, then refined with ICP, on 190 shape pairs of FAUST re-meshed\label{fig:geo curves for Table desc funmaps}.
}
\end{figure}

\section{Proof of Theorem~\ref{thm:discrete_Q}}
\label{app:proof_discrete_Q}
    Let $\varphi : M \rightarrow N$ be the permutation associated to the matrix $\Pi$.

    The energy of Eq.~\eqref{eq:best_Q} for vertex-based vector field must be evaluated for $X$ a basis of complex field and $f$ a basis of functions. We choose the ``hat" basis for the fields $X^i \in \C^{|V_M|}$ where $X^i_j = z \delta_{ij}, z = 1 \text{ or } \imath$ and $f^i \in \R^{|V_M|}$ where $f^i_j = \delta_{ij}$. The symbol $\delta$ denotes the Kronecker delta. Explicitly writing the coefficients of the matrices, boils down to:
    \begin{equation*}
        (C D_{X^i}^M)_{kj} = \left| \begin{matrix}
        \left\langle z , (\nabla f^j)_i \right\rangle , & k = \varphi(i) \\
        0 , & k \neq \varphi(i) \\
        \end{matrix} \right.
    \end{equation*}
    \begin{equation*}
        (D_{Q X^i}^N C)_{kj} = \left| \begin{array}{ll}
        \left\langle Q (X^i)_k , (\nabla f^{\varphi(j)})_k \right\rangle , & (k \varphi(j)) \in E_N \text{ or } k = \varphi(j) \\
        0 , & (k \varphi(j)) \notin E_N \text{ and } k \neq \varphi(j) \\
        \end{array} \right.
    \end{equation*}

    An immediate conclusion is that $Q$ must be zero everywhere except at $Q_{\varphi(i),i}$ so there exists vector $q \in \C^{|V_M|}$ such that:
    \begin{equation*}
        Q = D(q) \Pi .
    \end{equation*}

     Now we can go back the least-squares problem and find each coefficient of $q$ individually. Using the fact that $z$ form a basis of $\C$, at a vertex $i \in M$ the best conformal deformation of the tangent plane $q_{\varphi(i)}$ is solution of:
    \begin{equation*}
        q_{\varphi(i)} = \arg\min_{x \in \C} \sum_{j} \left| \left\langle x , (\nabla f^{\varphi(j)})_{\varphi(i)} \right\rangle - \left\langle 1 , (\nabla f^j)_i \right\rangle \right|^2 .
    \end{equation*}

\section{Proof of Theorem~\ref{thm:isometry_Q}}
\label{app:proof_isometry_Q}
    \begin{itemize}
        \item Suppose that $Q$ is an isometric pushforward then the commutativity with the Laplacian immediately holds as the connection Laplacian, like the usual cotangent Laplacian, is preserved by isometric changes. 
        \item Assume that $Q$ commutes with the Laplacian then thanks to Thm.~\ref{thm:discrete_Q}, the cotan-weights are preserved under the mapping. Therefore, the deformation is an isometry~\cite{gu2010discrete}.
    \end{itemize}

\section{Solving the least-squares problem to estimate $Q$}
\label{sec:solve_Q_appendix}

\begin{algorithm}[t!]
\caption{Complex Discrete Optimisation}
 \label{alg:discop_ZO_Q}
\begin{algorithmic}[1]
\State \textbf{Intput:} Manifold meshes $M$ and $N$
\State Initial pointwise maps $\Pi_{MN}$
\State \textbf{Output:} Refined maps $\Pi^{ref}_{MN}$
\State \textbf{Parameters:} The number of refinement steps $J$ 
\State An array $[k_j], j \in [1,J]$ with the (increasing) number of spectral coordinates to use at each refinement step
\State \textbf{Preprocessing:} Compute the Laplace Beltrami eigenbases $\Phi^M$ and $\Phi^N$ (used for function in spectral basis)

\boldnext
\State Compute the connection Laplacian complex eigenbases $\Psi^M$ and $\Psi^N$ (used for vector fields in spectral basis, see Section \ref{sec:Discr_reduced_basis})
\boldnext
\State Compute the differential operators $D_{\Phi^M_i}$ and $D_{\Phi^N_i}$ for $i \in [1, k_J]$ (used for estimating $Q$ from $C$, see Section \ref{sec:discrete_pipeline_Q})
\boldnext
\State Compute the reduced divergence Operators $\divg_M$ and $\divg_N$ (used for conversion from $Q$ to pointwise map, see Section \ref{sec:Qtop2p})
 \For{$k \in [k_1, ..., k_J]$}
    \State $\Phi_M = \Phi^M_{[1,k]}, \Phi_N = \Phi^N_{[1,k]}$
    
    \boldnext
    \State $Q_{NM} = \argmin_{Q \in \mathcal{O}(k)} \sum_{i=1}^k \|C_{NM} D_{\Psi^N_i} - D_{Q \Psi^N_i} C_{NM}\|_F^2$

    \boldnext
    \State 
    $\Pi_{MN} = \text{NNsearch}(\divg_N \Psi_N, \divg_M \Psi_M Q_{NM})$
 
    \State $C_{NM} = \Phi_M^\dagger \Pi_{MN} \Phi_N$
    \State 
    $\Pi_{MN} = \text{NNsearch}(\divg_N \Psi_N,  \divg_M \Psi_M Q_{NM})$
    
\EndFor
 
\end{algorithmic}
\end{algorithm}

Here we propose to complete Section \ref{sec:discrete_pipeline_Q} by giving one potential way to solve Eq.~\eqref{eq:best_Q} and Eq.~\eqref{eq:best_Q_procrustes} explicitly. To that end we first proceed to rewrite Eq.~\eqref{eq:discrete_CQ}, by switching from functional to vector field operator. More precisely, for a shape $M$, and $f \in L^2(M), X \in TM$, we define the linear operator $D_f \in TM \rightarrow L^2(M)$:
$$
D_f(X)_p = \langle X, \nabla f \rangle_{T_pM}
$$
With this operator we adopt the dual point of view from $D_X$. Indeed $\forall f \in L^2(M), X \in TM, D_f(X) = D_X(f)$. Consequently, the discretization is almost identical, except that since this operator takes vector fields as input, we choose to encode it as a complex operator  $\textbf{D}_f \in \C^{|V| \times |V|}$. Namely, $\textbf{D}_f$ is a diagonal matrix such that $(\textbf{D}_f)_{ii} = \overline{{\nabla f}_i}$. One can then retrieve $D_f(X)$ by taking the real part $Re(\textbf{D}_f X)$. It can also be noted that $Im(\textbf{D}_f X) = Re(\textbf{D}_f \cdot \imath X) = D_f \cdot \imath X$, so that the complex matrix $\textbf{D}_f$ also stores information for rotated gradients.

Let $M$ and $N$ be two manifolds. Switching from $D_X$ to $D_f$ in Eq.~\eqref{eq:discrete_CQ}, we get Eq.~\eqref{eq:discrete_CQ_withf}:

\begin{equation}
C_{MN} D_{f}^M = D^{N}_{C_{MN} f} Q_{MN}, \quad \forall ~f \in \R^{|V_M|} .
\label{eq:discrete_CQ_withf}
\end{equation}

We then discuss how to minimize, for a fixed input functional map $C$ and a family of smooth functions $f_i$ the following energy. Similarly to Eq.~\eqref{eq:energy_CQ}, $f_i$ is chosen to be the truncated eigenbasis of the Laplace-Beltrami operator.

\begin{equation}
    E_{q}(Q) = \sum_i \| C D_{f_i}^M - D_{C f_i}^N Q \|^2_F,
\label{eq:energy_CDf-DCfQ}
\end{equation}

\paragraph*{Regular problem}

In this first paragraph we ignore the constraint $QQ^\star = I$, which corresponds to Eq.~\eqref{eq:best_Q}.

Consider an input functional map $C$, and reference functions $f_i, i \in [1,k_f]$.
The minimum to Eq.~\eqref{eq:energy_CDf-DCfQ} can be written explicitly: by concatenating $C D_{f_i}, i \in [1,k_f]$ in a big matrix $A \in \C^{(|V_N| \times k_f) \times |V_M|}$, and $D_{C f_i}, i \in [1,k_f]$ in another matrix $B \in \C^{(|V_N| \times k_f) \times |V_N|}$, we can re-write Eq.~\eqref{eq:discrete_CQ_withf} as $Q = B^\dagger A$, where $B^\dagger$ is the Moore pseudo-inverse of $B$. In practice, we express these operators in a reduced spectral basis (See Section \ref{sec:Discr_reduced_basis}) before computing the pseudo-inverse, to improve computation time. Indeed in the reduced basis, operators $A$ and $B$ are respectively of size $k_N k_f \times k_M$ and $k_N k_f \times k_N$ if we choose to truncate the tangent vector field eigenbasis at $k_M$ on $M$ and at $k_N$ on $N$. This results in a $k_N \times k_M$ reduced complex operator for $Q$.

\paragraph*{Procrustes problem}

Considering the same setting and keeping the last notations as in last paragraph, minimizing Eq.~\eqref{eq:energy_CDf-DCfQ} with the constraint $QQ^\star = I$, which corresponds to Eq.~\eqref{eq:best_Q_procrustes}, is a Procrustes problem. As such, it boils down to a Singular Value Decomposition: writing $\Omega = B^\star A$, we use its SVD $\Omega = U \Sigma V$ to retrieve the orthogonal matrix $Q = UV$ minimizing Eq.~\eqref{eq:best_Q_procrustes}. As stated in the previous paragraph, one benefits from writing these equations in the reduced basis.

\begin{figure}
\begin{center}
    \begin{overpic}
[trim=0.0cm 0.0cm 0.0cm 0.0cm,clip,width=1\linewidth]{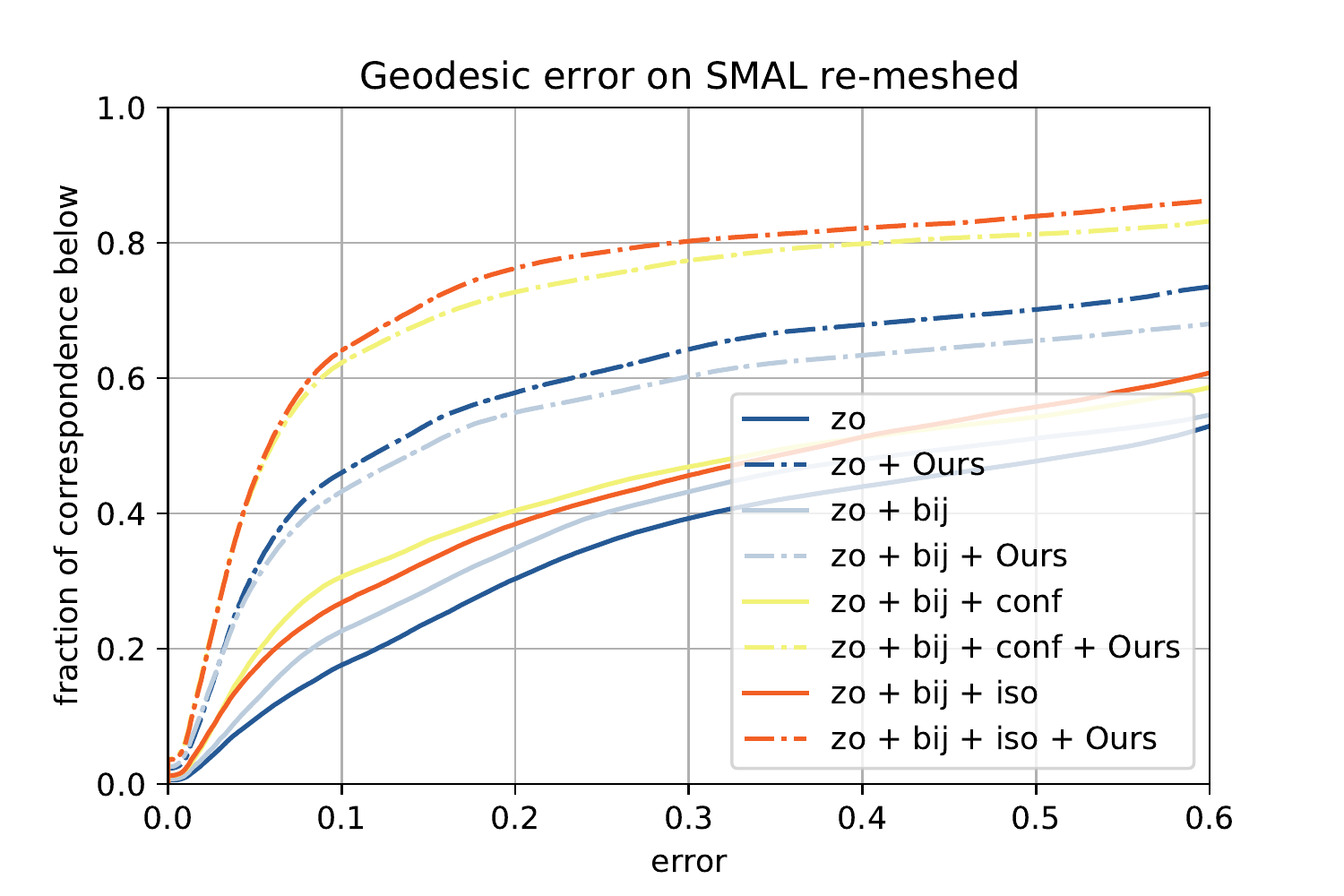}
\end{overpic}
\end{center}
\caption{Geodesic error of different versions of \textsc{ZoomOut} with and without our $Q$-step, on 50 shapes pairs of SMAL re-meshed\label{fig:geo curves SMAL}.
}
\end{figure}

\section{Vector Field transfer with more spectral values}
\label{app:VF_transfer}

In this appendix, we complete the vector field transfer experiment of Section \ref{sec:results_vf} by showing the evolution of vector field transfer accuracy with the number of spectral coordinates involved in both the Laplace-Beltrami and the connection Laplacian operator (the latter is only used in Azencot \textit{et al.} and our method). Table \ref{app_table:TVF transfer} reports this comparison in detail. Let us recall that only a number below $1$ shows a result with reasonable accuracy, above this threshold the error is higher that the norm of the input vector field.

We observe that the transfer from Wang \textit{et al.} is always very sensitive to noise, leading to high inaccuracies when noise corrupts the high frequencies. Notice also that the approach of Azencot \textit{et al.} only starts to give reasonable transfer with high spectral values, namely $k=150$. In comparison, our method almost doesn't suffer from noise, whether it is random or symmetric, and is accurate even for low spectral values. 

Our method is the only one to recover from strong symmetric noise ($a \geq 0.5$). Indeed the vector field we transfer for this second type of noise is antisymmetric as described in Section \ref{sec:results_vf}, but the input (blurred) functional map projects antisymmetric functions to $0$. This results in baselines transferring the input vector field to a vector field close to $0$ on the target shape, and thus errors above or close to $1$.

\begin{table}[t!]
\begin{center}
\begin{tabular}{|l|c|c|c|}
\hline
\multicolumn{4}{|c|}{Random noise} \\
\hline
Method / level of noise      &$s=0$&$s=0.2$&$s=0.5$\\
\hline
\multicolumn{4}{c}{$k=30$} \\
\hline
Wang \textit{et al.} \cite{wang2018vector}
& \textbf{0.42} & 5.3 & 14 \\
Azencot \textit{et al.} \cite{azencot2013operator}
& 11 & 12 & 12 \\
Ours & 0.54 &  \textbf{0.57}  & \textbf{0.80} \\
\hline
\multicolumn{4}{c}{$k=70$} \\
\hline
Wang \textit{et al.} \cite{wang2018vector}
& \textbf{0.41} & 14 & 33 \\
Azencot \textit{et al.} \cite{azencot2013operator}
& 2.6 & 2.4 & 2.4 \\
Ours & 0.46 &  \textbf{0.48}  & \textbf{0.77} \\
\hline
\multicolumn{4}{c}{$k=150$} \\
\hline
Wang \textit{et al.} \cite{wang2018vector}
& \textbf{0.37} & 27 & 68 \\
Azencot \textit{et al.} \cite{azencot2013operator}
& 0.68 & 0.81 & 1.0 \\
Ours & 0.44 &  \textbf{0.47}  & \textbf{0.81} \\
\hline
\hline
\multicolumn{4}{|c|}{Symmetric noise} \\
\hline
Method / level of noise      & $a=0.3$ & $a=0.5$ & $a=0.6$ \\
\hline
\multicolumn{4}{c}{$k=30$} \\
\hline
Wang \textit{et al.} \cite{wang2018vector}
& 0.87 & 1.13 & 1.29 \\
Azencot \textit{et al.} \cite{azencot2013operator}
& 10.19 & 6.84 & 10.59 \\
Ours & \textbf{0.54} &  \textbf{0.62}  & \textbf{0.76} \\
\hline
\multicolumn{4}{c}{$k=70$} \\
\hline
Wang \textit{et al.} \cite{wang2018vector}
& 0.81 & 1.09 & 1.26 \\
Azencot \textit{et al.} \cite{azencot2013operator}
& 2.93 & 1.99 & 2.55 \\
Ours & \textbf{0.39} &  \textbf{0.47}  & \textbf{0.81} \\
\hline
\multicolumn{4}{c}{$k=150$} \\
\hline
Wang \textit{et al.} \cite{wang2018vector}
& 0.77 & 1.05 & 1.22 \\
Azencot \textit{et al.} \cite{azencot2013operator}
& 0.69 & 1.04 & 1.43 \\
Ours & \textbf{0.37} &  \textbf{0.41}  & \textbf{0.56} \\
\hline
\end{tabular}
\end{center}
\caption{Average accuracy of the three vector field transfer algorithms of Section \ref{sec:results_vf} on $20$ random pairs of FAUST~\cite{FAUST} for two types of noise, \emph{and three noise levels}. We use $k=30, 70, 150$ eigenvectors for both real and complex Laplacian operators. 
}
\label{app_table:TVF transfer}
\end{table}

\section{Geodesic distance curves for Table \ref{table:funmap pipe with Q} and \ref{table:bij ZO with Q}}
\label{app:geo_curves}
In Figure \ref{fig:geo curves for Table desc funmaps} and \ref{fig:geo curves SMAL} we respectively display the results of Table \ref{table:funmap pipe with Q} and \ref{table:bij ZO with Q} with more precision using the Princeton graphs first introduced in \cite{kim2011blended}. With these curves it is easier to assert the finer quality of the correspondence computed with the help of complex functional maps.

\section{Complex bijective \zo{} algorithms}
\label{app:algo}

In this section we describe precisely how to implement the different versions of \zo{} to which we add our $Q$-step. These algorithms are used to generate the correspondence whose quality are reported in Figure \ref{fig:bijZOQ} and Table \ref{table:bij ZO with Q}.

\paragraph*{Complex bijective ZoomOut}

We first display the bijective \zo{} algorithm used in \cite{MapTree}, modified to include our $Q$-step. This results in Algorithm \ref{alg:bij_ZO_Q}, where we bold the line numbers where change occurs. The idea behind bijective \zo{} is to optimize for maps in both directions (from $M$ to $N$ \emph{and} from $N$ to $M$).
The energy to minimize is the so-called bijective energy $E_{bij} = \|C_{MN} C_{NM} - I\|^2_F + \|C_{NM} C_{MN} - I\|^2_F$ , and \cite{MapTree} proposes to optimize it with the following steps :
Remarking that $C_{NM} = \Phi_M^\dagger \Pi_{MN} \Phi_N$, we have $\|\Phi_M \|^2_F \|C_{NM} C_{MN} - I\|^2_F = \| \Pi_{MN} \Phi_N C_{MN} -  \Phi_M \|^2_F$. Thus, knowing $\Pi_{MN}$ and $\Pi_{NM}$, one can optimize for $C_{MN}$ in a bijective fashion with:
$$C_{MN} = \argmin_{C} \|\Phi_N C - \Pi_{NM} \Phi_M\|^2_F + \|\Pi_{MN} \Phi_N C - \Phi_M\|^2_F$$
And the same operation can be performed for $C_{NM}$.

Afterwards, one can perform a symmetric trick to get $\Pi_{MN}$ knowing $C_{MN}$ and $C_{NM}$, by additionally noting that in a bijective orthogonal setting, $C^{-1} = C^T$:
$$\Pi_{MN} = \argmin_{\Pi} \|\Pi \Phi_N C_{NM}^T- \Phi_M\|^2_F + \|\Pi \Phi_N C_{MN}- \Phi_M\|^2_F$$
And the same operation can be performed for $\Pi_{NM}$.

In the modified version of this algorithm, the pointwise map directly comes from our complex functional maps, and thus is orientation-aware. Its translation to functional map will also carry that information, making Algorithm \ref{alg:bij_ZO_Q} more robust than its original version to symmetry errors.


\paragraph*{Complex Discrete Optimization}

Secondly, we modify discrete optimisation algorithms \cite{rendiscrete21} (again, the necessary modifications bear bold line numbers). These algorithms consist in reducing a continuous energy with the same kind of trick used in bijective \zo. It is thus adaptable to all kind of energies. Besides, this method proved more efficient than continuous solvers to reduce natural energies such as conformality or isometry. one simply defines either $E(\Pi_{MN})$ (respectively $E(\Pi_{MN}, \Pi_{NM})$ in the case of a bijective energy), rewrites a continuous energy using the pointwise map trick $C_{MN} = \Phi_N^\dagger \Pi_{NM} \Phi_M$, and minimizes it for $\Pi_{MN}$ given $\C_{NM}$. Adding in our $Q$-step, we get Algorithm \ref{alg:discop_ZO_Q}. This algorithm is both orientation-aware \emph{and} optimizes for desirable maps such as isometries, often resulting in the best map as shown in Table \ref{table:bij ZO with Q} and Figure \ref{fig:bijZOQ}.

\end{document}